\documentclass[final]{cvpr}

\usepackage{times}
\usepackage{epsfig}
\usepackage{graphicx}
\usepackage{amsmath}
\usepackage{amssymb}
\usepackage{tabularx} 
\usepackage{array}
\usepackage{multirow}
\usepackage{booktabs}
\usepackage{soul}

\usepackage{cuted}
\usepackage{capt-of}

\usepackage[pagebackref=true,breaklinks=true,colorlinks,citecolor = green,bookmarks=false]{hyperref}
\newcommand\Autoref[1]{{\hypersetup{linkcolor=red}\autoref{#1}}}

\newcommand{\image}{\mathbf{I}}
\newcommand{\feature}{\mathbf{F}}
\newcommand{\dist}{d}
\newcommand{\geo}{g}
\newcommand{\p}{\mathbf{p}}
\newcommand{\q}{\mathbf{q}}

\newcommand{\corr}{\text{corr}}

\DeclareMathOperator*{\argmin}{arg\,min}

\def\cvprPaperID{2187} 
\def\confYear{CVPR 2021}

\begin{document}

\title{HumanGPS: Geodesic PreServing Feature for Dense Human Correspondences}

\author{Feitong Tan$^{1,2,}$\thanks{Work done while the author was an intern at Google.} \ \ Danhang Tang$^{1}$ \ \ Mingsong Dou$^{1}$ \ \ Kaiwen Guo$^{1}$ \ \ Rohit Pandey$^{1}$ \ \ Cem Keskin$^{1}$ \\ Ruofei Du$^{1}$ \ \ Deqing Sun$^{1}$ \ \ Sofien Bouaziz$^{1}$ \ \ Sean Fanello$^{1}$ \ \ Ping Tan$^{2}$ \ \ Yinda Zhang$^{1}$ \\
$^{1}$ Google  \ \ $^{2}$ Simon Fraser University \\ \\
}

\maketitle

\begin{strip}\centering
\vspace{-20mm}
\includegraphics[width=18cm]{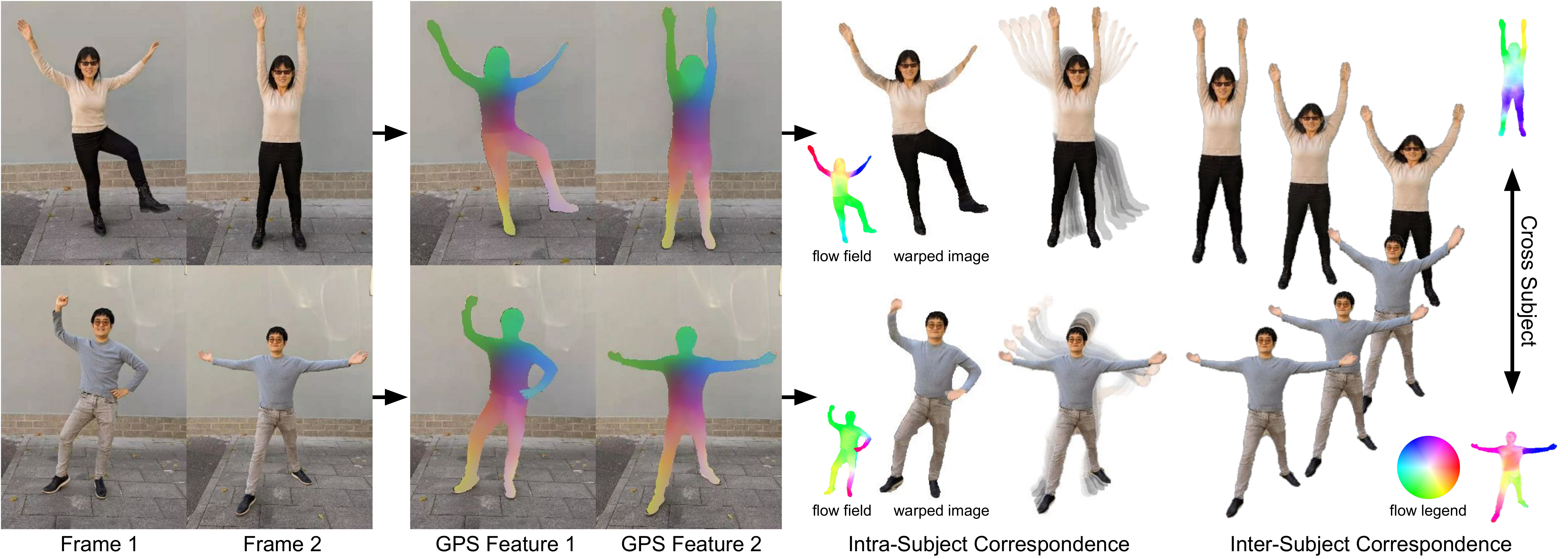}
\captionof{figure}{We propose a deep learning framework to learn a Geodesic PreServing (GPS) feature from RGB images that can produce accurate dense human correspondences. From left to right, we show the input images, extracted features visualized in color, the intra-subject correspondences, and the reconstruction of frame 1 using frame 2, the morphing \cite{liao2014semi,liao2014automating} between frames and across subjects, and the inter-subject correspondences.
Please refer to the color legend in the bottom right for the flow directions and magnitudes.}
\label{fig:teaser}
\end{strip}



\begin{abstract}
In this paper, we address the problem of building dense correspondences between human images under arbitrary camera viewpoints and body poses.
Prior art either assumes small motion between frames or relies on local descriptors, which cannot handle large motion or visually ambiguous body parts, e.g., left vs. right hand.
In contrast, we propose a deep learning framework that maps each pixel to a feature space, where the feature distances reflect the geodesic distances among pixels as if they were projected onto the surface of a 3D human scan.
To this end, we introduce novel loss functions to push features apart according to their geodesic distances on the surface. 
Without any semantic annotation, the proposed embeddings automatically learn to differentiate visually similar parts and align different subjects into an unified feature space.
Extensive experiments show that the learned embeddings can produce accurate correspondences between images with remarkable generalization capabilities on both intra and inter subjects.~\footnote{Project webpage:~\href{https://feitongt.github.io/HumanGPS/}{https://feitongt.github.io/HumanGPS/}}
\end{abstract}


\vspace{-6mm}
\section{Introduction}
Finding correspondences across images is one of the fundamental problems in computer vision and it has been studied for decades.
With the rapid development of digital human technology, building dense correspondences between human images has been found to be particularly useful for many applications, such as non-rigid tracking and reconstruction \cite{dou2016fusion4d,dou2017motion2fusion,newcombe2015dynamicfusion,du2019montage4d}, neural rendering \cite{tewari2020state}, and appearance transfer \cite{zanfir2018human,wu2019m2e}.
Traditional approaches in computer vision extract image features on local keypoints and generate correspondences between points with similar descriptors after performing a nearest neighbor search, {\em e.g.}, SIFT \cite{lowe1999object}.
More recently, deep learning methods \cite{long2014convnets,yi2016lift,schuster2019sdc,gaurmanifold}, replaced hand-crafted components with full end-to-end pipelines.
Despite their effectiveness on many tasks, these methods often deliver sub-optimal results when performing dense correspondences search on humans, due to the high variation in human poses and camera viewpoints and visual similarity between body parts.
As a result, the existing methods either produce sparse matches, {\em e.g.}, skeleton joints \cite{openpose}, or dense but imprecise correspondences \cite{guler2018densepose}.

\begin{figure}
\center
\includegraphics[width=8cm]{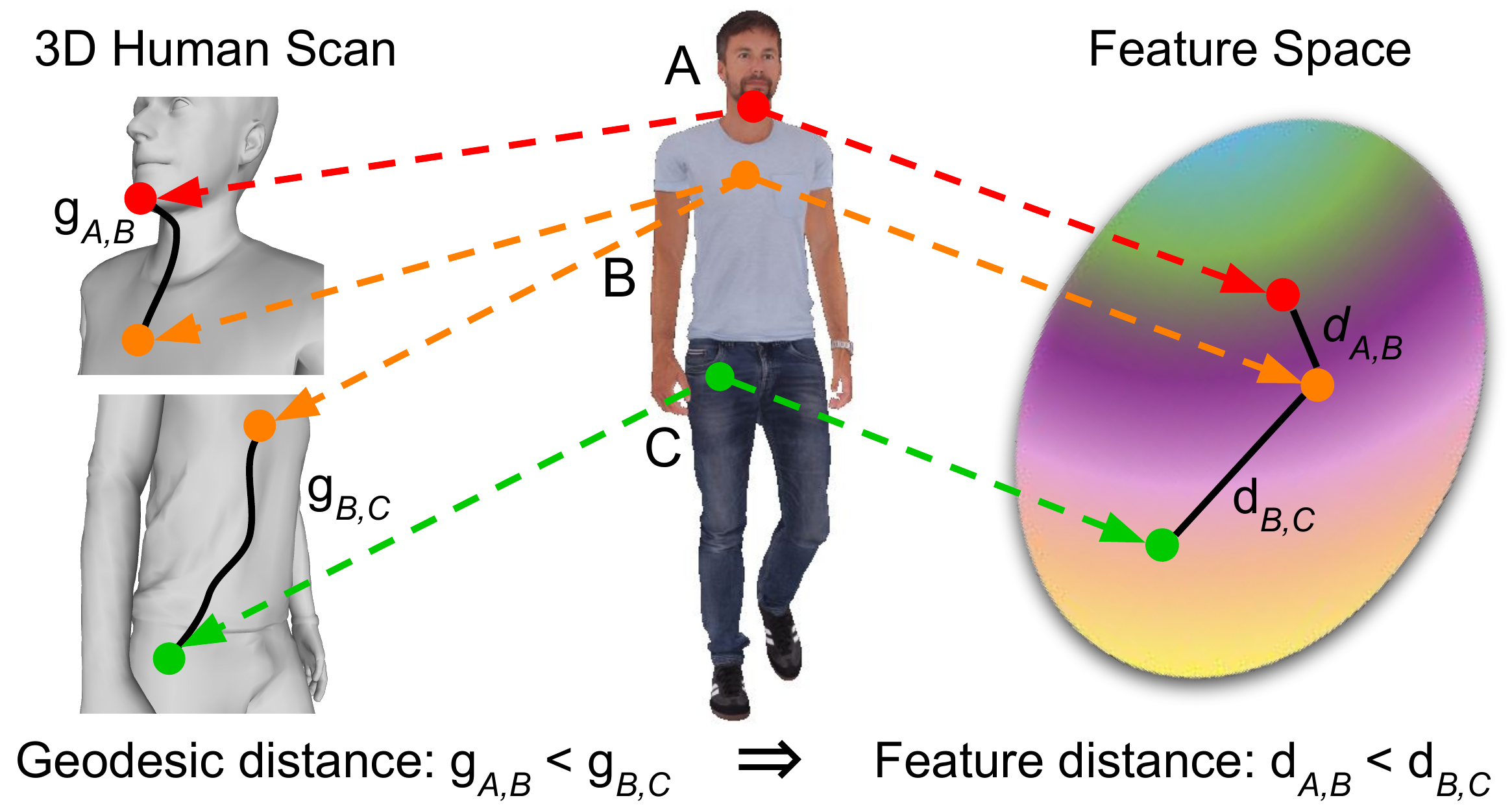}
\caption{Core idea: we learn a mapping from RGB pixels to a feature space that preserves geodesic properties of the underlying 3D surface. 
The 3D geometry is only used in the training phase.
}
\label{fig:core_idea}
\vspace{-15pt}
\end{figure}

In this paper, we propose a deep learning method to learn a \textbf{Geodesic PreServing (GPS)} feature taking RGB images as input, which can lead to accurate dense correspondences between human images through nearest neighbor search (see \Autoref{fig:teaser}).
Differently from previous methods using triplet loss \cite{schuster2019sdc,hoffer2015deep}, {\em i.e.} hard binary decisions, we advocate that the feature distance between pixels should be inversely correlated to their likelihood of being correspondences, which can be intuitively measured by the geodesic distance on the 3D surface of the human scan (\Autoref{fig:core_idea}).
For example, two pixels having zero geodesic distance means they project to the same point on the 3D surface and thus a match, and the probability of being correspondences becomes lower when they are apart from each other, leading to a larger geodesic distance.
While the geodesic preserving property has been studied in 3D shape analysis~\cite{shamai2017geodesic,kokkinos2012intrinsic,moreno2011deformation}, e.g., shape matching, we are the first to extend it for dense matching in image space, which encourages the feature space to be strongly correlated with an underlying 3D human model, and empirically leads to accurate, smooth, and robust results.

To generate supervised geodesic distances on the 3D surface, we leverage 3D assets such as RenderPeople \cite{renderpeople} and the data acquired with The Relightables \cite{guo2019relightables}. These high quality 3D models can be rigged and
allow us to generate pairs of rendered images from the same subject under different camera viewpoints and body poses, together with geodesic distances between any locations on the surface. In order to enforce soft, efficient, and differentiable constraints, we propose novel single-view and cross-view dense geodesic losses, where features are pushed apart from each other with a weight proportional to their geodesic distance.

We observe that the GPS features not only encode local image content, but they also have a strong semantic meaning. Indeed, even without any explicit semantic annotation or supervision, we find that our features automatically differentiate semantically different locations on the human surface and it is robust even in ambiguous regions of the human body ({\em e.g.}, left hand vs. right hand, torso vs. back). Moreover, we show that the learned features are consistent across different subjects, {\em i.e.}, the same semantic points from other persons still map to a similar feature, without any inter-subject correspondence data provided during the training.

In summary, we propose to learn an embedding that significantly improves the quality of the dense correspondences between human images. The core idea is to use the geodesic distance on the 3D surface as an effective supervision and combine it with novel loss functions to learn a discriminative feature. 
The learned embeddings are effective for dense correspondence search, and they show remarkable intra- and inter-subjects robustness without the need of any cross-subject annotation or supervision.
We show that our approach achieves state-of-the-art performance on both intra- and inter-subject correspondences and that the proposed framework can be used to boost many crucial computer vision tasks that rely on robust and accurate dense correspondences, such as optical flow, human dense pose regression \cite{guler2018densepose}, dynamic fusion \cite{newcombe2015dynamicfusion} and image-based morphing \cite{liao2014semi,liao2014automating}.

\section{Related Work}
In this section, we discuss current approaches in the literature for correspondence search tasks. 

\vspace{2mm}
\noindent\textbf{Hand-Crafted and Learned Descriptors.} 
Traditional approaches that tackle the matching problem between two or more images typically rely on feature descriptors \cite{lowe2004distinctive,surf,daisy,rublee2011orb} extracted on sparse keypoints, which nowadays are still popular for Structure-From-Motion or SLAM systems when computational budget is limited.
More recently, machine learning based feature extractors are proposed for image patches by pre-training on classification tasks \cite{long2014convnets}, making binary decision on pairs \cite{han2015matchnet,zagoruyko2015learning,patchCollider} or via a triplet loss \cite{yi2016lift,balntas2016pn,mishchuk2017working,tian2017l2,luo2018geodesc,hoffer2015deep}.
Recently, Schuster \etal \cite{schuster2019sdc} proposed an architecture with stacked dilated convolutions to increase the receptive field.
These methods are designed for generic domain and do not incorporate domain specific knowledge, \textit{e.g.}, human body in our case.
When the domain is given, a unified embedding can be learned to align objects within a category to enforce certain semantic properties \cite{gaurmanifold,Thewlisnips,Thewlis19a,wang2019normalized}. The resulting intra-domain correspondences are arguably better than previous approaches.
While most of methods are built purely on RGB images, 3D data are used to either as the input \cite{wei2016dense}, provide cycle consistency \cite{zhou2016learning}, or create normalized label space \cite{wang2019normalized}.
In contrast, our method is designed specifically for human correspondences, takes only color images as input, and enforces informative constraints from the 3D geometry according to the geodesic distance on the human surface.

\vspace{2mm}
\noindent\textbf{Direct Regression of Correspondences.}
Orthogonal approaches to correspondence search aim at regressing directly the matches between images. Examples of this trend are optical flow methods that can estimate dense correspondences in image pairs.
Early optical flow methods are often built with hand-crafted features and formulated as energy minimization problems based on photometric consistency and spatial smoothness~\cite{horn1981determining,brox2004high,wedel2009structure}.


More recently, deep learning methods have become popular in optical flow \cite{dosovitskiy2015flownet,hui2018liteflownet,hur2019iterative,teed2020raft} and stereo matching \cite{yi2018learning,chang2018pyramid,zbontar2016stereo}, where they aim at learning end-to-end correspondences directly from the data.
PWC-Net~\cite{sun2018pwc} and LiteFlowNet~\cite{hui2018liteflownet} incorporate ideas from traditional methods and present a popular design using a feature pyramid, warping and a cost volume. IRR-PWC~\cite{hur2019iterative} and RAFT~\cite{teed2020raft} present iterative residual refinements, which lead to state-of-the-art performance.
These methods aim at solving the generic correspondence search problem and are not designed specifically for human motion, which could be large and typically non-rigid.

\vspace{2mm}
\noindent\textbf{Human Correspondences.}
There are plenty of works studying sparse human correspondences by predicting body keypoints describing the human pose in images \cite{papandreou,papandreou2018personlab,openpose}.
For dense correspondences, many works rely on an underlying parametric model of a human, such as SMPL \cite{SMPL:2015}, and regress direct correspondences that lie on the 3D model. This 3D model shares the same topology across different people, hence allowing correspondences to be established \cite{xiang2019monocular,guler2018densepose,zhu2020simpose}.
DensePose \cite{guler2018densepose} is the first method showing that such correspondences are learnable given a sufficiently large training set, although it requires heavy labor to guarantee the labeling quantity and quality.
Follow up work reduces the annotation workload using equivariance \cite{neverova2019slim} or simulated data \cite{zhu2020simpose}, and extend DensePose to enable pose transfer across subjects \cite{denseposetransfer}.




Differently from previous approaches, we show how to learn human specific features directly from the data, without the need of explicit annotations.
Our approach learns an embedding from RGB images that follows the geodesic properties of an underlying 3D surface.
Thanks to this, the proposed method can be applied to full body images, performs robustly to viewpoint and pose changes, and surprisingly generalizes well across different people without using any inter-subject correspondences during the training.

\section{Method}
\label{sec:method}

\begin{figure*}
\center
\includegraphics[width=18cm]{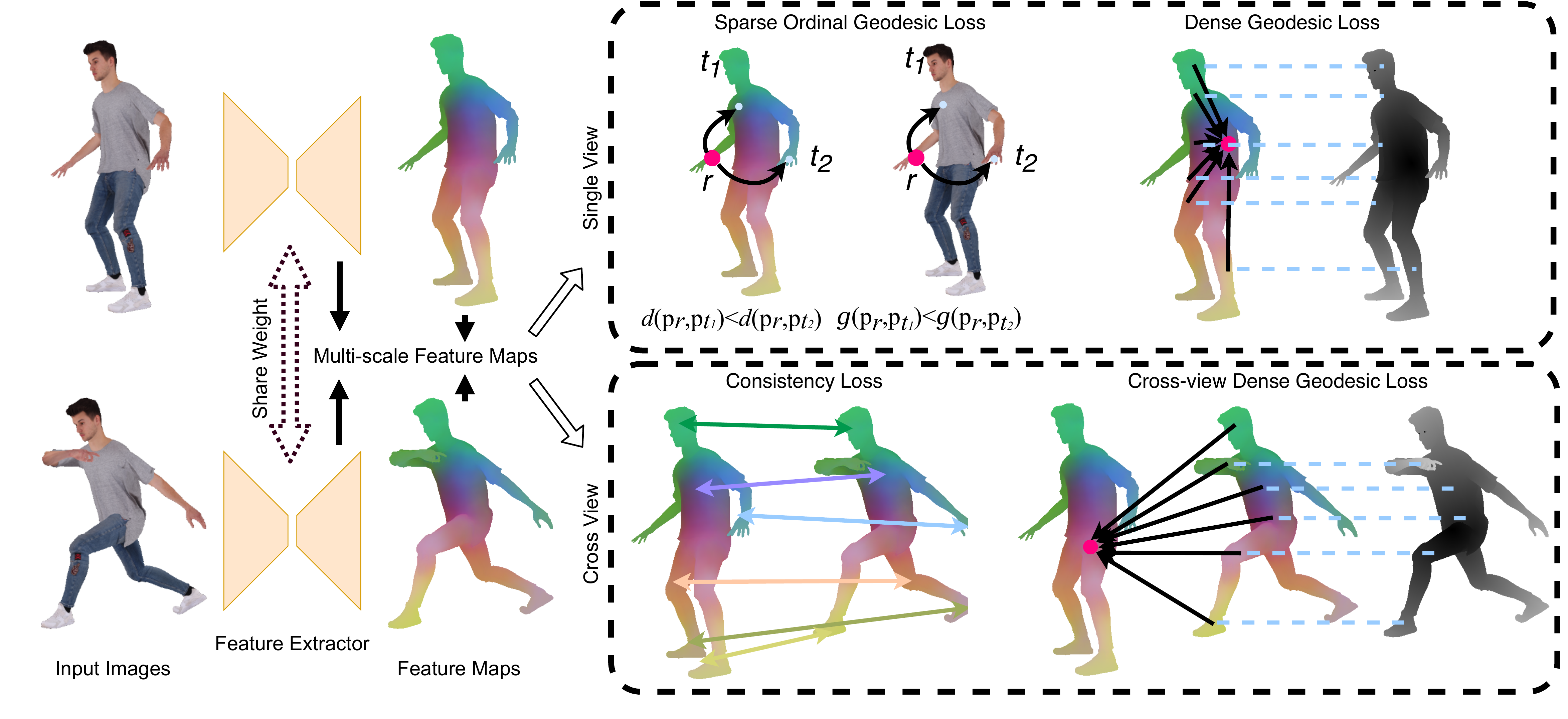}
\caption{Learning Human Geodesic PreServing Features. We train a neural network to extract features from RGB images. The learned embedding reflects the geodesic distance among pixels projected on the 3D surface of the human and can be used to build accurate dense correspondences. We train our feature extractor with a combination of consistency loss, sparse ordinal geodesic loss and dense intra/cross view geodesic loss: see text for details.}
\label{fig:pipeline}
\vspace{-15pt}
\end{figure*}

In this section, we introduce our deep learning method for dense human correspondences from RGB images (\Autoref{fig:pipeline}).
The key component of our method is a feature extractor, which is trained to produce a Geodesic PreServing (GPS) feature for each pixel, where the distance between descriptors reflects the geodesic distance on surfaces of the human scans.
We first explain the GPS feature in detail and then introduce our novel loss functions to exploit the geodesic signal as supervision for effective training. This enhances the discriminative power of the feature descriptors and reduces ambiguity for regions with similar textures.

\subsection{Geodesic PreServing Feature}
\label{sec:gps_explain}
Our algorithm starts with an image $\image \in \mathbb{R}^{H\times W \times 3}$ of height $H$ and width $W$, where we first run an off-the-shelf segmentation algorithm \cite{deeplabv3} to detect the person.
Then, our feature extractor takes as input this image and maps it into a high-dimensional feature map of the same spatial resolution $\feature \in \mathbb{R}^{H\times W \times C}$, where $C=16$ in our experiments.
The dense correspondences between two images $\image_1, \image_2$ can be built by searching for the nearest neighbor in the feature space, \ie, $\corr(\p)= \argmin_{\q\in \image_2}\dist(\p,\q), \forall \p \in \image_1$, where $\dist$ is a distance function defined in the feature space, and $\corr(\p)$ is the correspondence for the pixel $\p$ from $\image_1$ to $\image_2$.
In our approach, we constrain the feature for each pixel to be a unit vector $\|\feature_\image(\p)\|_2=1, \forall \p \in \image$ and use the cosine distance $\dist(\p,\q) =  1 - \feature_{\image_1}(\p)\cdot \feature_{\image_2}(\q)$.

Since images are 2D projections of the 3D world, ideally $\feature$ should be aware of the underlying 3D geometry of the human surface and be able to measure the likelihood of two pixels being a correspondence.
We find that the geodesic distance on a 3D surface is a good signal of supervision and thus should be preserved in the feature space, \ie, $\dist(\p, \q)\propto \geo(\p,\q), \forall \p,\q\in (\image_1, \image_2)$, where $\geo(\p,\q)$ is the geodesic distance between the projection of two pixels $\p,\q$ to 3D locations on the human surface (\Autoref{fig:core_idea}).

\vspace{2mm}
\noindent\textbf{Network Architecture.} Theoretically, any network architecture producing features in the same spatial resolution of the input image could be used as backbone for our feature extractor.
For the sake of simplicity, we utilize a typical 7-level U-Net~\cite{unet} with skip connections.
To improve the capacity without significantly increasing the model size, we add residual blocks inspired by~\cite{zhang2018road}.
More details can be found in supplementary materials.

\subsection{Loss Functions}
\label{sec:loss}
Our model uses a pair of images as a single training example. These images capture the same subjects under different camera viewpoints and body poses.
Both images are fed into the network and converted into feature maps. Multiple loss terms are then combined to compute the final loss function that is minimized during the training phase.

\vspace{2mm}
\noindent\textbf{Consistency Loss.} The first loss term minimizes the feature distance between ground truth corresponding pixels: $L_c(\p)=\sum \dist(\p, \corr(\p))$. 
Note however, that training with only $L_c$ will lead to degenerative case where all the pixels are mapped to the same feature.

\vspace{2mm}
\noindent\textbf{Sparse Ordinal Geodesic Loss.}
To prevent this degenerative case, previous methods use triplet loss \cite{schuster2019sdc,tripletloss} to increase the distance between non-matching pixels, \eg $\dist(\p, \corr(\p))\ll \dist(\p,\q), \forall \q \ne \corr(\p)$.
Whereas the general idea makes sense and works decently in practice, this loss function penalizes all the non-matching pixels equally without capturing their relative affinity, which leads to non-smooth and imprecise correspondences.

An effective measurement capturing the desired behavior is the geodesic distance on the 3D surface. This distance should be $0$ for corresponding pixels and gradually increase when two pixels are further apart.
To enforce a similar behavior in the feature space, we extend the triplet loss by randomly sampling a reference point $\p_r$ and two target points $\p_{t_1}, \p_{t_2}$, and defining a sparse ordinal geodesic loss:

\vspace{-3mm}
\begin{equation}
  L_s =
  \log(1+\exp(s\cdot(d(\p_r,\p_{t_1})-d(\p_r,\p_{t_2}))),
\end{equation}
where $s=\operatorname{sgn}(g(\p_r, \p_{t_2})-g(\p_r,\p_{t_1}))$. This term encourages the order between two pixels with respect to a reference pixel in feature space to be same as measured by the geodesic distance on the surface, and as a result, a pair of points physically apart on the surface tends to have larger distance in feature space.

\vspace{2mm}
\noindent\textbf{Dense Geodesic Loss.}
$L_s$ penalizes the order between a randomly selected pixels pair, which, however, does not produces an optimal GPS feature. One possible reason is due to the complexity of trying to order all the pixels, which is a harder task compared to the original binary classification method proposed for the triplet loss.
In theory, we could extend $L_s$ to order all the pixels in the image, which unfortunately is non-trivial to run efficiently during the training.

Instead, we relax the ordinal loss and define a dense version of the geodesic loss between one randomly picked pixel $\p_r$ to all the pixels $\p_t$ in the image:
\begin{equation}
  L_{d} = \sum_{\p_t\in \image}
   \log\left(1+\exp(g(\p_r,\p_t)-d(\p_r,\p_t)\right).  \\
\end{equation}

This loss, again, pushes features between non-matching pixels apart, depending on the geodesic distance.
It does not explicitly penalize the wrong order, but it is effective for training since all the pixels are involved in the loss function and contribute to the back-propagation.

\vspace{2mm}
\noindent\textbf{Cross-view Dense Geodesic Loss.}
The features learned with the aforementioned loss terms produces overall accurate correspondence, but susceptible to visually similar body parts.
For example in \Autoref{fig:activation_map} (top row), the feature always matches the wrong hand, since it does not capture correctly the semantic part due to the presence of large motion.
To mitigate this issue, we extend  $L_d$ and define it between pairs of images such that: given a pixel $\p_1$ on $\image_1$ and $\p_2$ on $\image_2$:

\begin{equation}
  L_{cd} = \sum_{\p_2 \in \image_2}
   \log\left(1+\exp(g(\corr(\p_1),\p_2)-d(\p_1,\p_2)\right).  \\
\end{equation}

At its core, the intuition behind this loss term is very similar $L_d$, except that it is cross-image and provides the network with training data with high variability due to viewpoint and pose changes.
We also tried to add a cross-view sparse ordinal geodesic loss but found it not improving.

\begin{figure*}[t]
\begin{center}
\includegraphics[width=\linewidth]{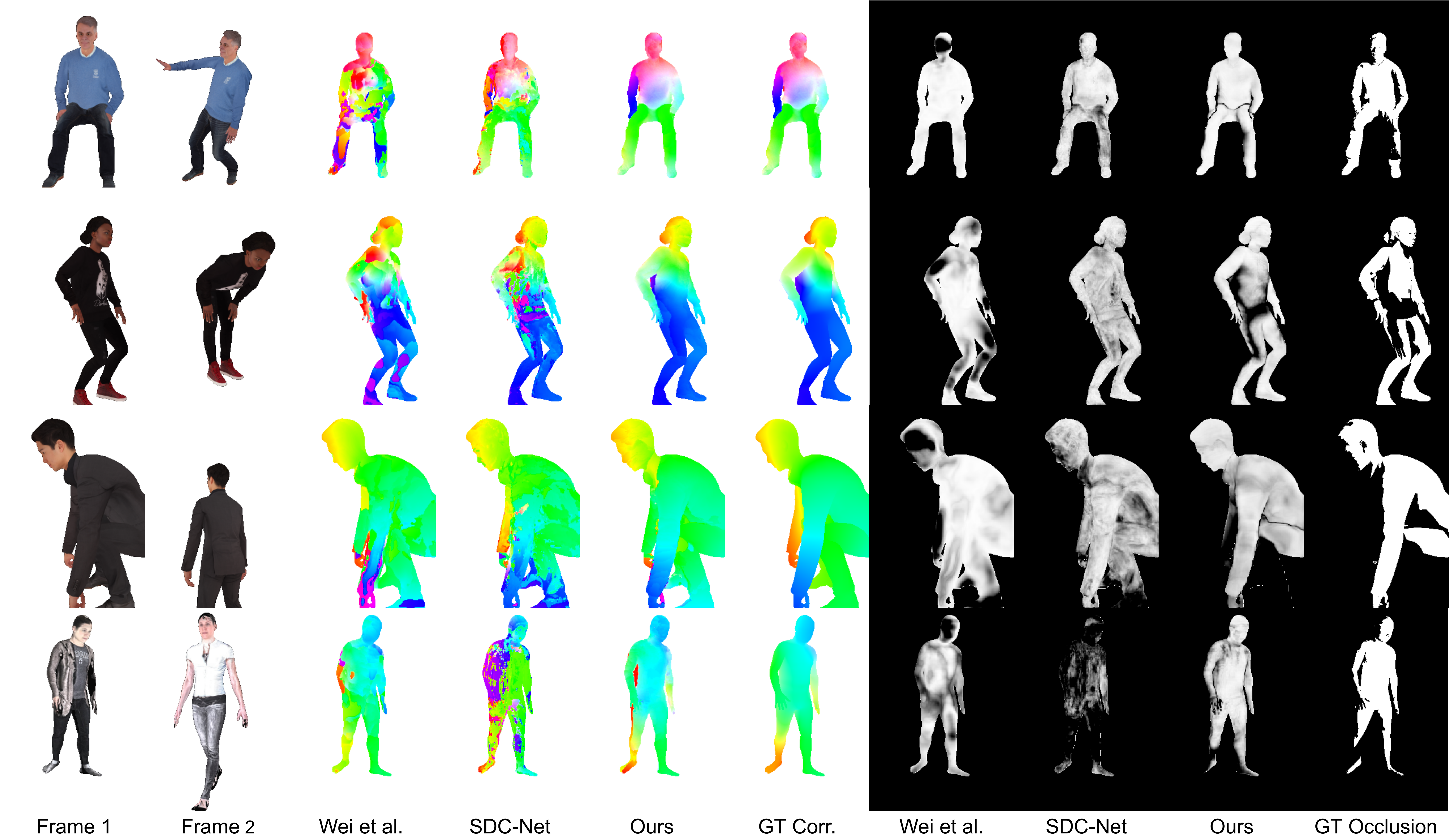}
\end{center}
\vspace{-0.15in}
\caption{Dense correspondences (visualized as optical flow) built via nearest neighbor search and the predicted visibility masks. Our results are more accurate, smooth, and free from obvious mistakes when compared to other methods. On the right, we show the visibility probability map obtained via the distance to the nearest neighbor. Note that our feature successfully captures occluded pixels ({\em i.e.}, dark pixels) in many challenging cases. The method is effective for both intra-subjects (rows 1-3) and inter-subjects (row 4).}
\label{fig:vis_search_flow}
\vspace{-0.1in}
\end{figure*}

\vspace{2mm}
\noindent\textbf{Total Loss.}
The total loss function is a weighted sum of the terms detailed above $L_t = w_cL_c+w_sL_s+w_dL_d+w_{cd}L_{cd}$. The weights are set to $1.0$, $3.0$, $5.0$, $3.0$ for $w_c, w_s, w_d, w_{cd}$ respectively. 
The weight for each term is chosen empirically such that the magnitude of gradients from each loss is roughly comparable.
To encourage robustness across different scales, we compute this loss at each intermediate level of the decoder, and down-weight the loss to $\frac{1}{8}$.
As demonstrated in our ablation studies, we found this increases the overall accuracy of the correspondences.

Our whole system is implemented in TensorFlow 2.0 \cite{abadi2016tensorflow}. The model is trained with batch size of 4 using ADAM optimizer. 
The learning rate is initialized at $1 \times 10^{-4}$ and reduces to $70\%$ for every 200K iterations.
The whole training takes 1.6 millions iterations to converge.

\begin{table*}[t]
\centering

\begin{tabular}{l||cc|cc|cc||cc}
\hline 
\multirow{3}{*}{\qquad \qquad Methods} & \multicolumn{6}{c||}{Intra-Subject} & \multicolumn{2}{c}{Inter-Subject}\tabularnewline
\cline{2-9}  
 & \multicolumn{2}{c|}{ SMPL \cite{loper2015smpl} } & \multicolumn{2}{c|}{Relightables \cite{guo2019relightables}} & \multicolumn{2}{c||}{RenderPeople \cite{renderpeople}} & \multicolumn{2}{c}{SMPL \cite{loper2015smpl}}\tabularnewline
\cline{2-9} 
 & non & all & non & all & non & all & non & all\tabularnewline
\hline 
SDC-Net \cite{schuster2019sdc} & 16.96 & 33.17 & 17.79 & 29.14 & 20.07 & 39.95 & 81.48 & 96.60\tabularnewline
Wei \etal  \cite{wei2016dense} & 18.08 & 30.59 & 29.54 & 43.64 & 34.42 & 46.23 & 18.03 & 31.55\tabularnewline
Ours + Full + Multi-scale & \textbf{7.12} & \textbf{17.51} & \textbf{11.24} & \textbf{18.95} & \textbf{11.91} & \textbf{22.12} & \textbf{8.49} & \textbf{17.99}\tabularnewline
\hline 
Ours + triplet & 9.14 & 24.34 & 13.18 & 25.59 & 16.84 & 29.80 & 21.08 & 30.75\tabularnewline
Ours + classify & 9.73 & 25.80 & 15.97 & 33.03 & 18.33 & 34.03 & 11.21 & 25.72\tabularnewline
Ours + $L_c$ + $L_s$   & 8.17 & 19.31 & 14.61 & 21.45 & 14.51 & 24.21 & 12.02 & 24.51\tabularnewline
Ours + $L_c$ + $L_s$ + $L_d$ & 7.50 & 18.00 & 12.24 & 19.30 & 12.41 & 22.73 & 9.19 & 18.61\tabularnewline
Ours + Full & 7.32 & 17.57 & 11.50 & 19.12 & 12.29 & 22.48 & 8.57 & \textbf{17.87}\tabularnewline
Ours + Full + Multi-scale & \textbf{7.12} & \textbf{17.51} & \textbf{11.24} & \textbf{18.95} & \textbf{11.91} & \textbf{22.12} & \textbf{8.49} & 17.99\tabularnewline
\hline 
\end{tabular}
\vspace{1mm}
\caption{Quantitative evaluation for correspondences search. We report the average end-point-error (EPE) of non-occluded (marked as non) and all pixels (marked as all) on four test sets created from different sources of 3D assets. Our model significantly outperforms previous methods for descriptor learning on all the datasets. The model trained with the proposed loss is better than all the ablation models trained with other alternatives. We report the results for both intra and inter-subjects.}
\label{tab:aepe}
\vspace{-0.15in}
\end{table*}

\section{Experiments}
In this section, we evaluate the GPS features using multiple datasets and settings. In particular, we compare our approach to other state-of-the-art methods for correspondence search and show its effectiveness for both intra- and inter-subjects problems.
Additional evaluations and applications, such as dynamic fusion \cite{newcombe2015dynamicfusion} and image-based morphing \cite{liao2014semi,liao2014automating}, can be found in the supplementary materials.

\subsection{Data Generation}
\label{sec:scans}
Since it would be very challenging to fit 3D geometry on real images, we resort to semi-synthetic data, where photogrammetry or volumetric capture systems are employed to capture subjects under multiple viewpoints and illumination conditions.
In particular, we generate synthetic renderings using SMPL~\cite{SMPL:2015}, RenderePeople~\cite{renderpeople}, and The Relightables~\cite{guo2019relightables}. These 3D assets are then used to obtain correspondences across views and geodesic distances on the surface. As demonstrated by the previous work~\cite{saito2020pifuhd,zhu2020simpose}, training on captured models generalizes well on real images in the wild. 
A similar idea has been used to create a fully synthetic dataset for optical flow~\cite{dosovitskiy2015flownet}, but in this work we focus on human images with larger camera viewpoints and body pose variations. All the details regarding the data generation can be found in the supplementary material.

\subsection{Dense Matching with Nearest Neighbor Search}
We first evaluate the capability of the GPS features in building dense pixel-wise matches via nearest neighbor search (\Autoref{sec:gps_explain}).

\vspace{2mm}
\noindent\textbf{Baseline Methods.}
We compare to two state-of-the-art descriptor learning methods that use different supervision for the learning. 1) SDC-Net~\cite{schuster2019sdc}: The SDC-Net extracts dense feature descriptors with stacked dilated convolutions, and is trained to distinguish correspondences and non-matching pixels by using threshold hinge triplet loss~\cite{bailer2017cnn}.
2) Wei~\etal~\cite{wei2016dense}: This method learns to extract dense features from a depth image via classification tasks on over-segmentations.
For fair comparison, we over segment our human models and rendered the segmentation label to generated the training data, i.e. over-segmentation label map. The network is trained with the same classification losses but takes color images as the input.

\vspace{2mm}
\noindent\textbf{Data and Evaluation Metrics.}
For each source of human scans introduced in~\Autoref{sec:scans}, we divide the 3D models into train and test splits, and render image pairs for training and testing respectively.
Additionally, we also generate a testing set using SMPL to quantitatively evaluate inter subjects performances since the SMPL model provides cross-subject alignment, which can be used to extract inter-subject correspondence ground truth.

As for evaluation metrics, we use the standard average end-point-error (AEPE) between image pairs, computed as the $\ell_2$ distance between the predicted and ground-truth correspondence pixels on the image.

\begin{figure}
\center
\includegraphics[width=8.5cm]{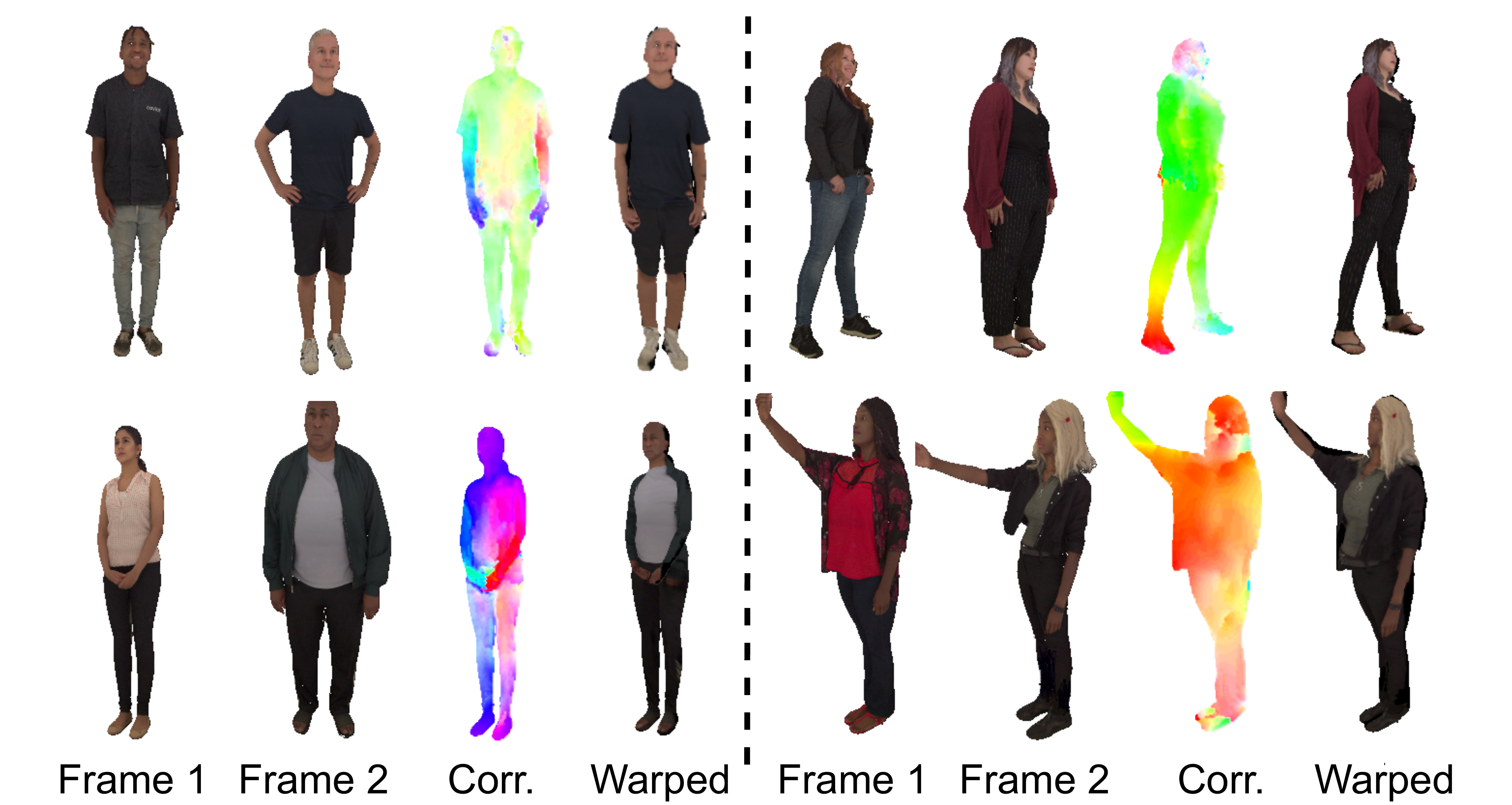}

\caption{Inter-subject correspondences and warp fields. Note how our approach correctly preserves the shape of the reference Frame 1 while plausibly warping the texture from Frame 2.}
\label{fig:vis_inter}
\vspace{-15pt}
\end{figure}


\vspace{2mm}
\noindent\textbf{Performance on Intra-Subject Data.}
We first evaluate our method for intra-subject correspondences. 
All the methods are re-trained on three training sets and tested on each test split respectively.
The dense correspondence accuracy of our approach and two state-of-art methods on the each test sets are shown in \Autoref{tab:aepe} (Intra-Subject).
Our method consistently achieves significantly lower error on both non-occluding and all pixels in all three datasets.

\Autoref{fig:vis_search_flow} (row 1-3) shows the qualitative results of the correspondences built between two images, visualized as flow where hue and saturation indicates the direction and magnitude (See \Autoref{fig:teaser} for color legend).
Our method generates much smooth and accurate correspondences compared to other methods and makes less mistakes for ambiguous body parts, {\em e.g.}, hands.


\paragraph{Performance on Inter-Subject Data.}
We also compare our approach on inter-subject data although no explicit cross-subject correspondences are provided during the training.
The results are shown in~\Autoref{tab:aepe} (Inter-Subject).
SDC-Net \cite{schuster2019sdc} does not generalize to inter-subject data since the triplet loss only differentiates positive and negative pairs and does not learn any human prior.
Wei~\etal~\cite{wei2016dense} shows some generalization thanks to the dense classification loss but the average end-point-error is much higher compared to our approach.
Comparatively, our method generalizes well to the inter-subject with error slightly higher, but roughly comparable to the intra-subject SMPL test set.
\Autoref{fig:vis_search_flow} (row 4) shows qualitative results of inter-subject correspondences. Again our method significantly outperforms other methods. In~\Autoref{fig:vis_inter} we show additional examples where we also build an image warp using the correspondences to map one person to another: notice how we preserve the shape of the body while producing a plausible texture. More results are presented in supplementary materials.

\paragraph{Occluded Pixels}
As mentioned in \Autoref{sec:method}, our feature space learns to preserve the surface geodesic distance, which is a measurement of the likelihood of the correspondence.
If a pixel cannot find a matching that is close enough in the feature space, it is likely that the pixel is not visible ({\em i.e.}, occluded) in the other view.
Inspired by this, we retrieve a visibility mask via the distance of each pixel to the nearest neighbor in the other image, {\em i.e.}, $d_{nn}= \min_{\q\in \image_2}\dist(\p,\q), \forall \p \in \image_1$.
 Specifically, the visibility is defined as $1-d_{nn}$ for SDC-Net and our method using cosine distance, and $\tfrac{1-d_{nn}}{2000}$ for Wei \etal using $\ell_2$ distance.
\Autoref{fig:vis_search_flow} (Right) visualizes the visibility map, {\em i.e.}, dark pixels are occluded.
Our method effectively detects occluded pixel more accurately than the other methods. More details can be found in supplementary material.

\begin{figure}
\center
\includegraphics[width=8cm]{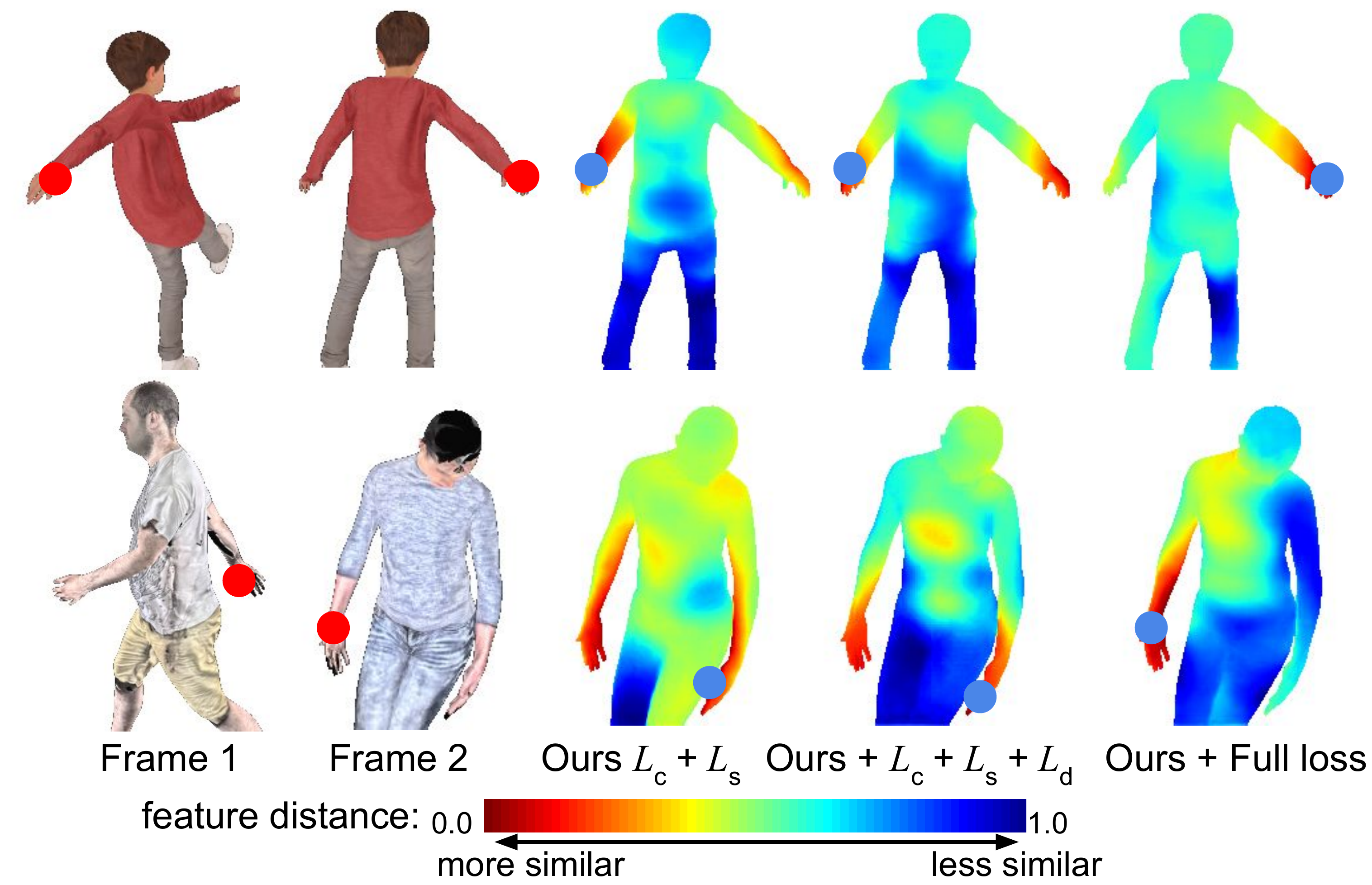}
\caption{Distance maps on intra- and inter- subjects, using models trained with different loss functions. We visualize the feature distance between the pixel in the left image (marked with a red dot) to all the pixels in the image on the right. The closest correspondence is marked with a blue dot. Our full loss provides the best feature space with less ambiguity, {\em e.g.}, between left and right hand. }
\label{fig:activation_map}
\vspace{-15pt}
\end{figure}

\begin{table*}[t]
\centering

\vspace{-3mm}
\begin{tabular}{l||cc|cc|cc||cc}
\hline 
\multirow{3}{*}{\qquad Methods} & \multicolumn{6}{c||}{Intra-Subject} & \multicolumn{2}{c}{Inter-Subject}\tabularnewline
\cline{2-9} 
 & \multicolumn{2}{c|}{\quad SMPL \cite{loper2015smpl} \quad} & \multicolumn{2}{c|}{Relightables \cite{guo2019relightables}} & \multicolumn{2}{c||}{RenderPeople \cite{renderpeople}} & \multicolumn{2}{c}{SMPL \cite{loper2015smpl}}\tabularnewline
\cline{2-9}  
 & non & all & non & all & non & all & non & all\tabularnewline
\hline 
PWC-Net \cite{sun2018pwc} & 4.51 & 13.27 & 3.57 & 10.01 & 9.57 & 18.74 & 20.06 & 28.14\tabularnewline
PWC-Net* & 3.89 & 12.82 & 3.42 & 9.39 & 7.91 & 16.99 & 19.21 & 23.77\tabularnewline
PWC-Net + GPS & \textbf{2.73} & \textbf{10.86} & \textbf{2.99} & \textbf{9.01} & \textbf{6.89} & \textbf{14.72} & \textbf{12.08} & \textbf{17.92}\tabularnewline
\hline 
RAFT \cite{tf-raft}& 3.62 & 12.30 & 3.27 & 11.65 & 6.74 & 15.90 & 45.47 & 53.09\tabularnewline
RAFT* & 3.24 & 12.82 & 2.79 & 11.39 & 5.62 & 14.79 & 57.82 & 66.04\tabularnewline
RAFT + GPS   & \textbf{2.13} & \textbf{10.12} & \textbf{2.27} & \textbf{10.52} & \textbf{3.95} & \textbf{12.68} & \textbf{10.76} & \textbf{17.61}\tabularnewline
\hline 
\end{tabular}

\vspace{1mm}
\caption{Quantitative evaluation on dense human correspondences via SoTA optical flow networks - PWC-Net \cite{sun2018pwc} and RAFT \cite{teed2020raft}. 
On both architecture, integrating with our GPS feature achieve the best performance. See text for the $^*$ models. 
}
\label{tab:optical_flow_aepe}
\vspace{-15pt}
\end{table*}

\subsection{Ablation Study}
In this section, we study the effect of each loss term for learning the GPS features.
We add $L_s, L_d, L_{cd}$ gradually into the training and show the performance of correspondences through nearest neighbor search in \Autoref{tab:aepe} (bottom half).
For reference, we also train our feature extractor with losses from SDC-Net~\cite{schuster2019sdc} and Wei~\etal \cite{wei2016dense} for a strict comparison on the loss only (see ``Ours+triplet'' and ``Ours+classify'').
Training with our loss is more effective than the baseline methods, and the error on all the test sets, both intra- and inter-subject, keeps reducing with new loss terms added in.
This indicates that all the loss terms contributes the learning of GPS feature, which is consistent with our analysis (\Autoref{sec:loss}) that the loss terms improve the embeddings from different aspects.

To further analyse the effect of each loss term, we visualize the feature distance from one pixel in frame 1 (marked by a red dot) to all the pixels in frame 2 (red dot is the ground-truth correspondence) in \Autoref{fig:activation_map}.
The blue dots show the pixel with lowest feature distance, {\em i.e.}, the predicted matching.
Training with $L_s$ does not produce clean and indicative distance maps.
Adding $L_d$ improves the performance, but the feature is still ambiguous between visually similar parts, {\em e.g.}, left and right hand.
In contrast, the feature trained with all the losses produces the best distance map, and multi-scale supervision further improves the correspondence accuracy as shown in~\Autoref{tab:aepe} (last row).




\begin{figure}
\center
\includegraphics[width=\linewidth]{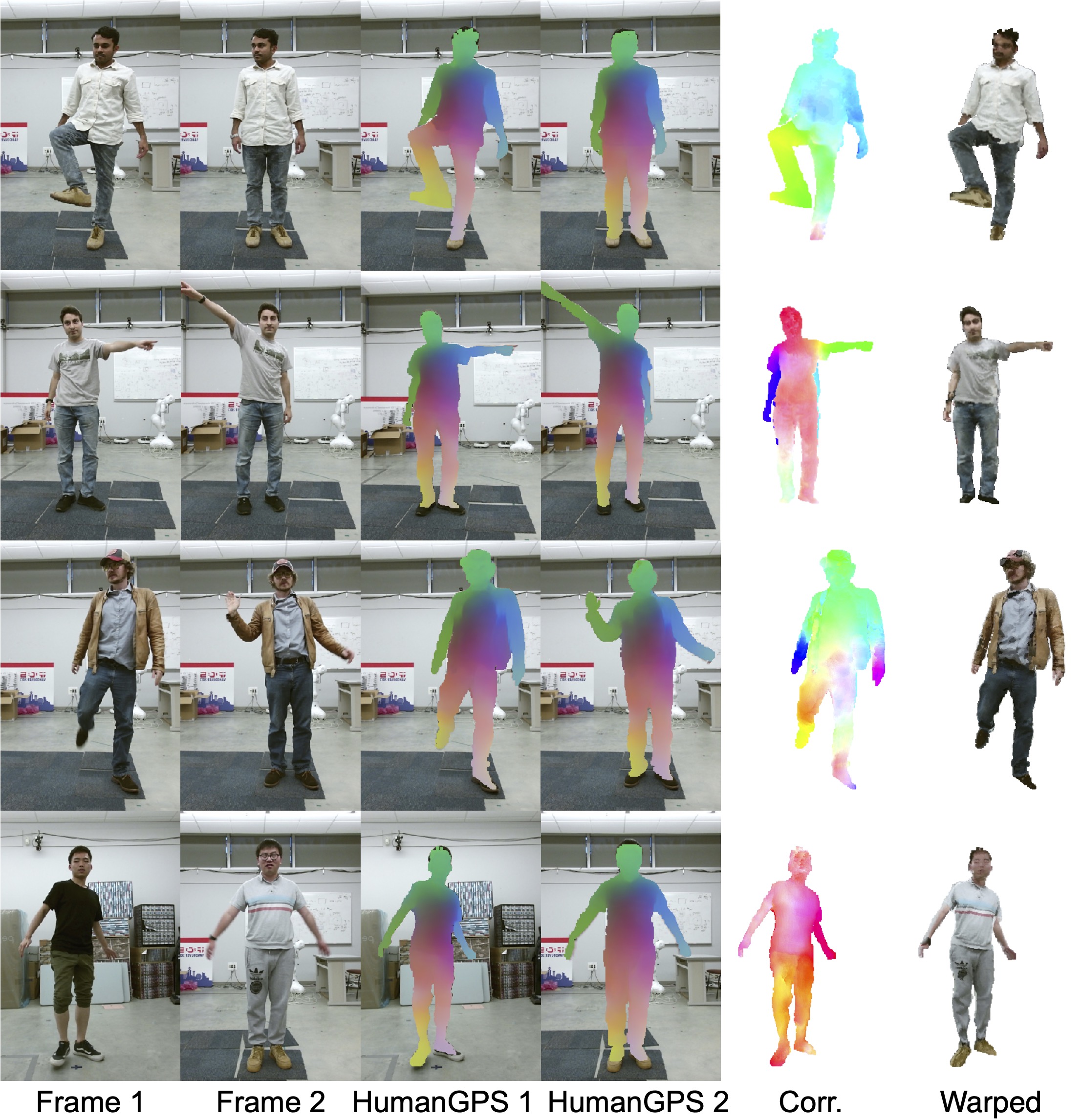}

\caption{Results on real images. HumanGPS generalizes in the wild and provides accurate correspondences across subjects. The correspondences (column 5) successfully warp frame 2 to frame 1 for both intra- and inter-subject cases (column 6).}
\label{fig:vis_real}
\vspace{-15pt}

\end{figure}

\subsection{Generalization to Real Images}
\label{sec:realimg}
Our model is fully trained on semi-synthetic data acquired with high-end volumetric capture systems ({\em e.g.}, RenderPeople~\cite{renderpeople} and The Relightables~\cite{guo2019relightables}), which help to minimize the domain gap. In this section, we assess how our method performs on real data.
To our knowledge, there is no real dataset with ground-truth dense correspondences for evaluation. Thus, we compute the cycle consistency across frames on videos in the wild. Specifically, given three frames from a video, we calculate correspondences among them,  cascade matching $\mathbf{I}_1$$\rightarrow$$\mathbf{I}_2$ and $\mathbf{I}_2$$\rightarrow$$\mathbf{I}_3$  and measure the average end-point-error (AEPE) 
to the ones calculated directly with $\mathbf{I}_1$$\rightarrow$$\mathbf{I}_3$. We collected 10 videos of moving performers. The avg. cycle consistency errors of SDC-Net, Wei~\etal and ours are $17.53$, $22.19$, and $12.21$ respectively. Also, we show the qualitative results on real images from~\cite{tang2019neural}.
As shown in \Autoref{fig:vis_real}, our method generalizes reasonably to the real data, producing an accurate feature space, correspondences, as well as warped images using the correspondences. For additional results and evaluation on annotated sparse keypoints, please see~\Autoref{fig:teaser} and the supplementary material.




\subsection{HumanGPS with End-to-end Networks}
In this section, we show that our HumanGPS features can be integrated with state-of-art end-to-end network architectures to improve various tasks.

\vspace{2mm}
\noindent\textbf{Integration with Optical Flow.}
We integrate our features to the state-of-the-art optical flow methods PWC-Net~\cite{sun2018pwc} and RAFT~\cite{teed2020raft}.
Both methods consist of a siamese feature extractor and a follow-up network to regress the flow.
We attach an additional feature extractor and pre-train it with our losses to learn our GPS features. The GPS features are then combined with the output of the original feature extractor by element-wise average and then fed into the remaining layers of the network to directly regress a 2D flow vector pointing to the matching pixel in the other image.

The quantitative evaluation is presented in~\Autoref{tab:optical_flow_aepe}. All the methods are trained on our training sets. To compare under the same model capacity, we also train both methods with the additional feature extractor but without our loss functions (marked with $^*$). Our GPS features benefits the correspondence learning on both SOTA architectures consistently over all test sets.
We also train PWC-Net to output an additional layer to predict the occlusion mask, and again the model trained with GPS features consistently outperforms all the baselines (see supplementary materials).

\vspace{2mm}
\noindent\textbf{Integration with DensePose.}
Our GPS features automatically maps features extracted from different subjects to the same embedding, and thus can benefit cross-subject tasks like dense human pose regression~\cite{guler2018densepose}.
To show this, we pre-train a GPS feature extractor on our datasets, attach two more MLP layers, and fine-tune on DensePose-COCO dataset \cite{guler2018densepose} to regress the UV coordinates.
To make sure the model capacity is comparable with other methods, we use the backbone of previous methods for feature extraction, {\em i.e.}, a ResNet-101 FCN in DensePose \cite{guler2018densepose}, and Hourglass in Slim DensePose \cite{neverova2019slim}.
To focus on the matching accuracy, we adopt their same evaluation setup, where ground truth bounding box is given; percentages of pixels with geodesic error less than certain thresholds are taken as the metric; and evaluate on DensePose MSCOCO benchmark \cite{guler2018densepose}.

\Autoref{tab:geodesic_error} shows the comparison with previous work.
Following DensePose~\cite{guler2018densepose}, we also train our model with their network backbone on full image and only foreground (marked as *).
Our method consistently achieves better performance than previous methods on various setting with different network backbones.
Note that the UV coordinates are estimated via only two layers of MLP, which indicates that the GPS features are already effective at mapping different subjects to the same embedding.
Please see supplementary material for qualitative results.

\begin{table}[t]
\begin{tabular}{llll}
\hline
\multicolumn{1}{c}{Methods} & \multicolumn{3}{c}{Accuracy}                                                                   \\ \cline{2-4}
\multicolumn{1}{c}{} & \multicolumn{1}{c}{5 cm} & \multicolumn{1}{c}{10 cm}&  \multicolumn{1}{c}{20 cm}     \\ \hline
DP ResNet-101 FCN \cite{guler2018densepose} & \multicolumn{1}{l}{43.05} & \multicolumn{1}{l}{65.23} & 74.17  \\
DP ResNet-101 FCN* \cite{guler2018densepose} & \multicolumn{1}{l}{51.32} & \multicolumn{1}{l}{75.50} & 85.76   \\
SlimDP HG - 1 stack \cite{neverova2019slim} & \multicolumn{1}{l}{49.89} & \multicolumn{1}{l}{74.04} & 82.98\\
\hline
Our ResNet-101 FCN  & \multicolumn{1}{l}{49.09} & \multicolumn{1}{l}{73.12} & 84.51  \\ 
Our ResNet-101 FCN* & \multicolumn{1}{l}{53.01} & \multicolumn{1}{l}{76.77} & 87.70  \\ 
Our HG - 1 stack  & \multicolumn{1}{l}{50.50} & \multicolumn{1}{l}{75.57} & 87.18  \\ \hline
\end{tabular}
\vspace{1mm}
\caption{Quantitative evaluation for dense human pose regression on DensePose COCO dataset~\cite{guler2018densepose}. Following~\cite{guler2018densepose}, we assume ground-truth bounding box is given and calculate percentage of pixels with error smaller than thresholds. We also compare models trained on full image and only foreground (marked by $^*$).}
\label{tab:geodesic_error}
\vspace{-15pt}
\end{table}

\section{Conclusion}

We presented a deep learning approach to build HumanGPS, a robust feature extractor for finding human correspondences between images. The learned embedding is enforced to follow the geodesic distances on an underlying 3D surface representing the human shape. By proposing novel loss terms, we show that the feature space is able to map human body parts to features that preserve their semantic. Hence, the method can be applied to both intra- and inter-subject correspondences. We demonstrate the effectiveness of HumanGPS via comprehensive experimental results including comparison with the SOTA methods, ablation studies, and generalization studies on real images.
In future work, we will extend HumanGPS to work with more object categories and remove the dependency of a foreground segmentation step.


{\small
\bibliographystyle{ieee_fullname}
\bibliography{humangps}

\begin{thebibliography}{10}\itemsep=-1pt

\bibitem{CMUMoCap}
Carnegie-mellon mocap database.
\newblock \url{base. http://mocap.cs.cmu.edu}.

\bibitem{Mixamo}
Mixamo.
\newblock \url{https://www.mixamo.com}.

\bibitem{renderpeople}
Renderpeople.
\newblock \url{https://renderpeople.com/}.

\bibitem{abadi2016tensorflow}
Mart{\'\i}n Abadi, Ashish Agarwal, Paul Barham, Eugene Brevdo, Zhifeng Chen,
  Craig Citro, Greg~S Corrado, Andy Davis, Jeffrey Dean, Matthieu Devin, et~al.
\newblock Tensorflow: Large-scale machine learning on heterogeneous distributed
  systems.
\newblock {\em arXiv preprint arXiv:1603.04467}, 2016.

\bibitem{bailer2017cnn}
Christian Bailer, Kiran Varanasi, and Didier Stricker.
\newblock Cnn-based patch matching for optical flow with thresholded hinge
  embedding loss.
\newblock In {\em Proceedings of the IEEE Conference on Computer Vision and
  Pattern Recognition}, pages 3250--3259, 2017.

\bibitem{balntas2016pn}
Vassileios Balntas, Edward Johns, Lilian Tang, and Krystian Mikolajczyk.
\newblock Pn-net: Conjoined triple deep network for learning local image
  descriptors.
\newblock {\em arXiv preprint arXiv:1601.05030}, 2016.

\bibitem{surf}
Herbert Bay, Andreas Ess, Tinne Tuytelaars, and Luc Van~Gool.
\newblock Speeded-up robust features (surf).
\newblock In {\em CVIU}, 2008.

\bibitem{bogo2016keep}
Federica Bogo, Angjoo Kanazawa, Christoph Lassner, Peter Gehler, Javier Romero,
  and Michael~J Black.
\newblock Keep it smpl: Automatic estimation of 3d human pose and shape from a
  single image.
\newblock In {\em European Conference on Computer Vision}, pages 561--578.
  Springer, 2016.

\bibitem{brox2004high}
Thomas Brox, Andr{\'e}s Bruhn, Nils Papenberg, and Joachim Weickert.
\newblock High accuracy optical flow estimation based on a theory for warping.
\newblock In {\em European conference on computer vision}, pages 25--36.
  Springer, 2004.

\bibitem{openpose}
Zhe Cao, Gines Hidalgo, Tomas Simon, Shih-En Wei, and Yaser Sheikh.
\newblock Openpose: realtime multi-person 2d pose estimation using part
  affinity fields.
\newblock {\em IEEE transactions on pattern analysis and machine intelligence},
  43(1):172--186, 2019.

\bibitem{chang2018pyramid}
Jia-Ren Chang and Yong-Sheng Chen.
\newblock Pyramid stereo matching network.
\newblock In {\em IEEE Conference on Computer Vision and Pattern Recognition
  (CVPR)}, 2018.

\bibitem{deeplabv3}
Liang-Chieh Chen, Yukun Zhu, George Papandreou, Florian Schroff, and Hartwig
  Adam.
\newblock Encoder-decoder with atrous separable convolution for semantic image
  segmentation.
\newblock {\em CoRR}, abs\/1802.02611, 2018.

\bibitem{dosovitskiy2015flownet}
Alexey Dosovitskiy, Philipp Fischer, Eddy Ilg, Philip Hausser, Caner Hazirbas,
  Vladimir Golkov, Patrick Van Der~Smagt, Daniel Cremers, and Thomas Brox.
\newblock Flownet: Learning optical flow with convolutional networks.
\newblock In {\em Proceedings of the IEEE International Conference on Computer
  Vision}, pages 2758--2766, 2015.

\bibitem{dou2017motion2fusion}
Mingsong Dou, Philip Davidson, Sean~Ryan Fanello, Sameh Khamis, Adarsh Kowdle,
  Christoph Rhemann, Vladimir Tankovich, and Shahram Izadi.
\newblock Motion2fusion: Real-time volumetric performance capture.
\newblock {\em ACM Transactions on Graphics (TOG)}, 36(6):1--16, 2017.

\bibitem{dou2016fusion4d}
Mingsong Dou, Sameh Khamis, Yury Degtyarev, Philip Davidson, Sean~Ryan Fanello,
  Adarsh Kowdle, Sergio~Orts Escolano, Christoph Rhemann, David Kim, Jonathan
  Taylor, et~al.
\newblock Fusion4d: Real-time performance capture of challenging scenes.
\newblock {\em ACM Transactions on Graphics (TOG)}, 35(4):1--13, 2016.

\bibitem{du2019montage4d}
Ruofei Du, Ming Chuang, Wayne Chang, Hugues Hoppe, and Amitabh Varshney.
\newblock {Montage4D: Real-Time Seamless Fusion and Stylization of Multiview
  Video Textures}.
\newblock {\em Journal of Computer Graphics Techniques}, 8(1):1--34, Jan. 2019.

\bibitem{gaurmanifold}
Utkarsh Gaur and BS Manjunath.
\newblock Weakly supervised manifold learning for dense semantic object
  correspondence.
\newblock In {\em Proceedings of the IEEE International Conference on Computer
  Vision}, pages 1735--1743, 2017.

\bibitem{girshick2014rich}
Ross Girshick, Jeff Donahue, Trevor Darrell, and Jitendra Malik.
\newblock Rich feature hierarchies for accurate object detection and semantic
  segmentation.
\newblock In {\em Proceedings of the IEEE Conference on Computer Vision and
  Pattern Recognition}, pages 580--587, 2014.

\bibitem{guler2018densepose}
R{\i}za~Alp G{\"u}ler, Natalia Neverova, and Iasonas Kokkinos.
\newblock Densepose: Dense human pose estimation in the wild.
\newblock In {\em CVPR}, 2018.

\bibitem{guo2019relightables}
Kaiwen Guo, Peter Lincoln, Philip Davidson, Jay Busch, Xueming Yu, Matt Whalen,
  Geoff Harvey, Sergio Orts-Escolano, Rohit Pandey, Jason Dourgarian, et~al.
\newblock The relightables: Volumetric performance capture of humans with
  realistic relighting.
\newblock {\em ACM Transactions on Graphics (TOG)}, 38(6):1--19, 2019.

\bibitem{twinfusion}
Kaiwen Guo, Jon Taylor, Sean Fanello, Andrea Tagliasacchi, Mingsong Dou, Philip
  Davidson, Adarsh Kowdle, and Shahram Izadi.
\newblock Twinfusion: High framerate non-rigid fusion through fast
  correspondence tracking.
\newblock In {\em 3DV}, 2018.

\bibitem{han2015matchnet}
Xufeng Han, Thomas Leung, Yangqing Jia, Rahul Sukthankar, and Alexander~C Berg.
\newblock Matchnet: Unifying feature and metric learning for patch-based
  matching.
\newblock In {\em Proceedings of the IEEE Conference on Computer Vision and
  Pattern Recognition}, pages 3279--3286, 2015.

\bibitem{hoffer2015deep}
Elad Hoffer and Nir Ailon.
\newblock Deep metric learning using triplet network.
\newblock In {\em International Workshop on Similarity-Based Pattern
  Recognition}, pages 84--92. Springer, 2015.

\bibitem{tripletloss}
Elad Hoffer and Nir Ailon.
\newblock Deep metric learning using triplet network.
\newblock In {\em ICLR (Workshop)}, 2015.

\bibitem{horn1981determining}
Berthold~KP Horn and Brian~G Schunck.
\newblock Determining optical flow.
\newblock In {\em Techniques and Applications of Image Understanding}, volume
  281, pages 319--331. International Society for Optics and Photonics, 1981.

\bibitem{hui2018liteflownet}
Tak-Wai Hui, Xiaoou Tang, and Chen Change~Loy.
\newblock Liteflownet: A lightweight convolutional neural network for optical
  flow estimation.
\newblock In {\em Proceedings of the IEEE Conference on Computer Vision and
  Pattern Recognition}, pages 8981--8989, 2018.

\bibitem{hur2019iterative}
Junhwa Hur and Stefan Roth.
\newblock Iterative residual refinement for joint optical flow and occlusion
  estimation.
\newblock In {\em Proceedings of the IEEE Conference on Computer Vision and
  Pattern Recognition}, pages 5754--5763, 2019.

\bibitem{ionescu2013human3}
Catalin Ionescu, Dragos Papava, Vlad Olaru, and Cristian Sminchisescu.
\newblock Human3. 6m: Large scale datasets and predictive methods for 3d human
  sensing in natural environments.
\newblock {\em IEEE transactions on pattern analysis and machine intelligence},
  36(7):1325--1339, 2013.

\bibitem{kokkinos2012intrinsic}
Iasonas Kokkinos, Michael~M Bronstein, Roee Litman, and Alex~M Bronstein.
\newblock Intrinsic shape context descriptors for deformable shapes.
\newblock In {\em 2012 IEEE Conference on Computer Vision and Pattern
  Recognition}, pages 159--166. IEEE, 2012.

\bibitem{need4speed}
Adarsh Kowdle, Christoph Rhemann, Sean Fanello, Andrea Tagliasacchi, Jon
  Taylor, Philip Davidson, Mingsong Dou, Kaiwen Guo, Cem Keskin, Sameh Khamis,
  David Kim, Danhang Tang, Vladimir Tankovich, Julien Valentin, and Shahram
  Izadi.
\newblock The need 4 speed in real-time dense visual tracking.
\newblock {\em SIGGRAPH Asia}, 2018.

\bibitem{liao2014semi}
Jing Liao, Rodolfo~S Lima, Diego Nehab, Hugues Hoppe, and Pedro~V Sander.
\newblock Semi-automated video morphing.
\newblock In {\em Computer Graphics Forum}, volume~33, pages 51--60. Wiley
  Online Library, 2014.

\bibitem{liao2014automating}
Jing Liao, Rodolfo~S Lima, Diego Nehab, Hugues Hoppe, Pedro~V Sander, and
  Jinhui Yu.
\newblock Automating image morphing using structural similarity on a halfway
  domain.
\newblock {\em ACM Transactions on Graphics (TOG)}, 33(5):1--12, 2014.

\bibitem{long2014convnets}
Jonathan~L Long, Ning Zhang, and Trevor Darrell.
\newblock Do convnets learn correspondence?
\newblock In {\em Advances in neural information processing systems}, pages
  1601--1609, 2014.

\bibitem{loper2014mosh}
Matthew Loper, Naureen Mahmood, and Michael~J Black.
\newblock Mosh: Motion and shape capture from sparse markers.
\newblock {\em ACM Transactions on Graphics (TOG)}, 33(6):1--13, 2014.

\bibitem{SMPL:2015}
Matthew Loper, Naureen Mahmood, Javier Romero, Gerard Pons-Moll, and Michael~J.
  Black.
\newblock {SMPL}: A skinned multi-person linear model.
\newblock {\em SIGGRAPH Asia}, 2015.

\bibitem{loper2015smpl}
Matthew Loper, Naureen Mahmood, Javier Romero, Gerard Pons-Moll, and Michael~J
  Black.
\newblock Smpl: A skinned multi-person linear model.
\newblock {\em ACM Transactions on Graphics (TOG)}, 34(6):1--16, 2015.

\bibitem{lowe1999object}
David~G Lowe.
\newblock Object recognition from local scale-invariant features.
\newblock In {\em Proceedings of the Seventh IEEE International Conference on
  Computer Vision}, volume~2, pages 1150--1157. Ieee, 1999.

\bibitem{lowe2004distinctive}
David~G Lowe.
\newblock Distinctive image features from scale-invariant keypoints.
\newblock {\em International journal of computer vision}, 60(2):91--110, 2004.

\bibitem{luo2018geodesc}
Zixin Luo, Tianwei Shen, Lei Zhou, Siyu Zhu, Runze Zhang, Yao Yao, Tian Fang,
  and Long Quan.
\newblock Geodesc: Learning local descriptors by integrating geometry
  constraints.
\newblock In {\em Proceedings of the European conference on computer vision
  (ECCV)}, pages 168--183, 2018.

\bibitem{mishchuk2017working}
Anastasiia Mishchuk, Dmytro Mishkin, Filip Radenovic, and Jiri Matas.
\newblock Working hard to know your neighbor's margins: Local descriptor
  learning loss.
\newblock In {\em Advances in Neural Information Processing Systems}, pages
  4826--4837, 2017.

\bibitem{mitchell1987discrete}
Joseph~SB Mitchell, David~M Mount, and Christos~H Papadimitriou.
\newblock The discrete geodesic problem.
\newblock {\em SIAM Journal on Computing}, 16(4):647--668, 1987.

\bibitem{moreno2011deformation}
Francesc Moreno-Noguer.
\newblock Deformation and illumination invariant feature point descriptor.
\newblock In {\em CVPR 2011}, pages 1593--1600. IEEE, 2011.

\bibitem{denseposetransfer}
Natalia Neverova, Riza~Alp G{\"{u}}ler, and Iasonas Kokkinos.
\newblock Dense pose transfer.
\newblock {\em ECCV}, 2018.

\bibitem{neverova2019slim}
Natalia Neverova, James Thewlis, Rıza~Alp Güler, Iasonas Kokkinos, and Andrea
  Vedaldi.
\newblock Slim densepose: Thrifty learning from sparse annotations and motion
  cues.
\newblock In {\em CVPR}, 2019.

\bibitem{newcombe2015dynamicfusion}
Richard~A Newcombe, Dieter Fox, and Steven~M Seitz.
\newblock Dynamicfusion: Reconstruction and tracking of non-rigid scenes in
  real-time.
\newblock In {\em Proceedings of the IEEE Conference on Computer Vision and
  Pattern Recognition}, pages 343--352, 2015.

\bibitem{papandreou2018personlab}
George Papandreou, Tyler Zhu, Liang-Chieh Chen, Spyros Gidaris, Jonathan
  Tompson, and Kevin Murphy.
\newblock Personlab: Person pose estimation and instance segmentation with a
  bottom-up, part-based, geometric embedding model.
\newblock In {\em ECCV}, 2018.

\bibitem{papandreou}
George Papandreou, Tyler Zhu, Nori Kanazawa, Alexander Toshev, Jonathan
  Tompson, Chris Bregler, and Kevin Murphy.
\newblock Towards accurate multi-person pose estimation in the wild.
\newblock In {\em CVPR}, 2017.

\bibitem{ranjan2020learning}
Anurag Ranjan, David~T Hoffmann, Dimitrios Tzionas, Siyu Tang, Javier Romero,
  and Michael~J Black.
\newblock Learning multi-human optical flow.
\newblock {\em International Journal of Computer Vision}, pages 1--18, 2020.

\bibitem{humanflow}
Anurag Ranjan, Javier Romero, and Michael~J. Black.
\newblock Learning human optical flow.
\newblock In {\em 29th British Machine Vision Conference}, Sept. 2018.

\bibitem{robinette2002civilian}
Kathleen~M Robinette, Sherri Blackwell, Hein Daanen, Mark Boehmer, and Scott
  Fleming.
\newblock Civilian american and european surface anthropometry resource
  (caesar), final report. volume 1. summary.
\newblock Technical report, SYTRONICS INC DAYTON OH, 2002.

\bibitem{unet}
Olaf Ronneberger, Philipp Fischer, and Thomas Brox.
\newblock U-net: Convolutional networks for biomedical image segmentation.
\newblock {\em MICCAI}, 2015.

\bibitem{rublee2011orb}
Ethan Rublee, Vincent Rabaud, Kurt Konolige, and Gary Bradski.
\newblock Orb: An efficient alternative to sift or surf.
\newblock In {\em 2011 International Conference on Computer Vision}, pages
  2564--2571. Ieee, 2011.

\bibitem{saito2020pifuhd}
Shunsuke Saito, Tomas Simon, Jason Saragih, and Hanbyul Joo.
\newblock Pifuhd: Multi-level pixel-aligned implicit function for
  high-resolution 3d human digitization.
\newblock In {\em Proceedings of the IEEE Conference on Computer Vision and
  Pattern Recognition}, June 2020.

\bibitem{schuster2019sdc}
Ren{\'e} Schuster, Oliver Wasenmuller, Christian Unger, and Didier Stricker.
\newblock Sdc-stacked dilated convolution: A unified descriptor network for
  dense matching tasks.
\newblock In {\em Proceedings of the IEEE Conference on Computer Vision and
  Pattern Recognition}, pages 2556--2565, 2019.

\bibitem{shamai2017geodesic}
Gil Shamai and Ron Kimmel.
\newblock Geodesic distance descriptors.
\newblock In {\em Proceedings of the IEEE Conference on Computer Vision and
  Pattern Recognition}, pages 6410--6418, 2017.

\bibitem{tf-raft}
Deqing Sun, Charles Herrmann, Varun Jampani, Michael Krainin, Forrester Cole,
  Austin Stone, Rico Jonschkowski, Ramin Zabih, William~T. Freeman, and Ce Liu.
\newblock {TF-RAFT}: A tensorflow implementation of raft.
\newblock In {\em ECCV Robust Vision Challenge Workshop}, 2020.

\bibitem{sun2018pwc}
Deqing Sun, Xiaodong Yang, Ming-Yu Liu, and Jan Kautz.
\newblock Pwc-net: Cnns for optical flow using pyramid, warping, and cost
  volume.
\newblock In {\em Proceedings of the IEEE Conference on Computer Vision and
  Pattern Recognition}, pages 8934--8943, 2018.

\bibitem{tang2019neural}
Sicong Tang, Feitong Tan, Kelvin Cheng, Zhaoyang Li, Siyu Zhu, and Ping Tan.
\newblock A neural network for detailed human depth estimation from a single
  image.
\newblock In {\em Proceedings of the IEEE International Conference on Computer
  Vision}, pages 7750--7759, 2019.

\bibitem{teed2020raft}
Zachary Teed and Jia Deng.
\newblock Raft: Recurrent all-pairs field transforms for optical flow.
\newblock In {\em European Conference on Computer Vision}, pages 402--419.
  Springer, 2020.

\bibitem{tewari2020state}
Ayush Tewari, Ohad Fried, Justus Thies, Vincent Sitzmann, Stephen Lombardi,
  Kalyan Sunkavalli, Ricardo Martin-Brualla, Tomas Simon, Jason Saragih,
  Matthias Nießner, Rohit Pandey, Sean Fanello, Gordon Wetzstein, Jun-Yan Zhu,
  Christian Theobalt, Maneesh Agrawala, Eli Shechtman, Dan~B Goldman, and
  Michael Zollhoefer.
\newblock State of the art on neural rendering.
\newblock In {\em Eurographics}, 2020.

\bibitem{Thewlis19a}
James Thewlis, Samuel Albanie, Hakan Bilen, and Andrea Vedaldi.
\newblock Unsupervised learning of landmarks by descriptor vector exchange.
\newblock In {\em International Conference on Computer Vision}.

\bibitem{Thewlisnips}
James Thewlis, Hakan Bilen, and Andrea Vedaldi.
\newblock Unsupervised learning of object frames by dense equivariant image
  labelling.
\newblock In {\em NIPS}. 2017.

\bibitem{tian2017l2}
Yurun Tian, Bin Fan, and Fuchao Wu.
\newblock L2-net: Deep learning of discriminative patch descriptor in euclidean
  space.
\newblock In {\em Proceedings of the IEEE Conference on Computer Vision and
  Pattern Recognition}, pages 661--669, 2017.

\bibitem{daisy}
Engin Tola, Vincent Lepetit, and Pascal Fua.
\newblock Daisy: An efficient dense descriptor applied to wide-baseline stereo.
\newblock In {\em PAMI}, 2010.

\bibitem{varol2017learning}
Gul Varol, Javier Romero, Xavier Martin, Naureen Mahmood, Michael~J Black, Ivan
  Laptev, and Cordelia Schmid.
\newblock Learning from synthetic humans.
\newblock In {\em Proceedings of the IEEE Conference on Computer Vision and
  Pattern Recognition}, pages 109--117, 2017.

\bibitem{wang2019normalized}
He Wang, Srinath Sridhar, Jingwei Huang, Julien Valentin, Shuran Song, and
  Leonidas~J. Guibas.
\newblock Normalized object coordinate space for category-level 6d object pose
  and size estimation.
\newblock In {\em CVPR}, 2019.

\bibitem{patchCollider}
Shenlong Wang, Sean~Ryan Fanello, Christoph Rhemann, Shahram Izadi, and
  Pushmeet Kohli.
\newblock The global patch collider.
\newblock {\em CVPR}, 2016.

\bibitem{wedel2009structure}
Andreas Wedel, Daniel Cremers, Thomas Pock, and Horst Bischof.
\newblock Structure-and motion-adaptive regularization for high accuracy optic
  flow.
\newblock In {\em 2009 IEEE 12th International Conference on Computer Vision},
  pages 1663--1668. IEEE, 2009.

\bibitem{wei2016dense}
Lingyu Wei, Qixing Huang, Duygu Ceylan, Etienne Vouga, and Hao Li.
\newblock Dense human body correspondences using convolutional networks.
\newblock In {\em Proceedings of the IEEE Conference on Computer Vision and
  Pattern Recognition}, pages 1544--1553, 2016.

\bibitem{wu2019m2e}
Zhonghua Wu, Guosheng Lin, Qingyi Tao, and Jianfei Cai.
\newblock M2e-try on net: Fashion from model to everyone.
\newblock In {\em Proceedings of the 27th ACM International Conference on
  Multimedia}, pages 293--301, 2019.

\bibitem{xiang2019monocular}
Donglai Xiang, Hanbyul Joo, and Yaser Sheikh.
\newblock Monocular total capture: Posing face, body, and hands in the wild.
\newblock In {\em Proceedings of the IEEE Conference on Computer Vision and
  Pattern Recognition}, 2019.

\bibitem{yi2016lift}
Kwang~Moo Yi, Eduard Trulls, Vincent Lepetit, and Pascal Fua.
\newblock Lift: Learned invariant feature transform.
\newblock In {\em ECCV}, 2016.

\bibitem{yi2018learning}
Kwang~Moo Yi*, Eduard Trulls*, Yuki Ono, Vincent Lepetit, Mathieu Salzmann, and
  Pascal Fua.
\newblock Learning to find good correspondences.
\newblock In {\em Proceedings of the IEEE Conference on Computer Vision and
  Pattern Recognition}, 2018.

\bibitem{zagoruyko2015learning}
Sergey Zagoruyko and Nikos Komodakis.
\newblock Learning to compare image patches via convolutional neural networks.
\newblock In {\em Proceedings of the IEEE Conference on Computer Vision and
  Pattern Recognition}, pages 4353--4361, 2015.

\bibitem{zanfir2018human}
Mihai Zanfir, Alin-Ionut Popa, Andrei Zanfir, and Cristian Sminchisescu.
\newblock Human appearance transfer.
\newblock In {\em Proceedings of the IEEE Conference on Computer Vision and
  Pattern Recognition}, pages 5391--5399, 2018.

\bibitem{zbontar2016stereo}
Jure Zbontar and Yann LeCun.
\newblock Stereo matching by training a convolutional neural network to compare
  image patches.
\newblock {\em Journal of Machine Learning Research (JMLR)}, 2016.

\bibitem{zhang2018road}
Zhengxin Zhang, Qingjie Liu, and Yunhong Wang.
\newblock Road extraction by deep residual u-net.
\newblock {\em IEEE Geoscience and Remote Sensing Letters}, 15(5):749--753,
  2018.

\bibitem{zhou2016learning}
Tinghui Zhou, Philipp Krahenbuhl, Mathieu Aubry, Qixing Huang, and Alexei~A
  Efros.
\newblock Learning dense correspondence via 3d-guided cycle consistency.
\newblock In {\em Proceedings of the IEEE Conference on Computer Vision and
  Pattern Recognition}, pages 117--126, 2016.

\bibitem{zhu2020simpose}
Tyler Zhu, Per Karlsson, and Christoph Bregler.
\newblock Simpose: Effectively learning densepose and surface normals of people
  from simulated data.
\newblock In {\em European Conference on Computer Vision}, pages 225--242.
  Springer, 2020.

\end{thebibliography}
}

\clearpage

\appendix{

\section*{Supplementary}

In this supplementary material, we provide details about our semi-synthetic dataset, our network architecture to learn the human geodesic preserving feature space, and additional experimental results.
Please see the additional webpage for video demos.

\section{Semi-synthetic Data}
As described in the main paper Section 4.1, our method relies on high quality 3D assets for training. To ensure high diversity and variation in our training set, we created synthetic datasets merging multiple state-of-art acquisition systems. 
Start from a 3D human scan, we render a pair of images from two different camera viewpoint and ground truths measurements, including 2D correspondences, geodesic distance between pixels and visibility masks.
Examples of generated data are shown in \Autoref{fig:dataset}.
In each example, we show the input pair of images, the ground truth correspondences with visibility mask, and the geodesic distance map w.r.t. one pixel (marked in red).
Row 1-3 shows intra-subject data, and Row 4 shows inter subject data.

\begin{figure*}
    \centering
    \includegraphics[width=\linewidth]{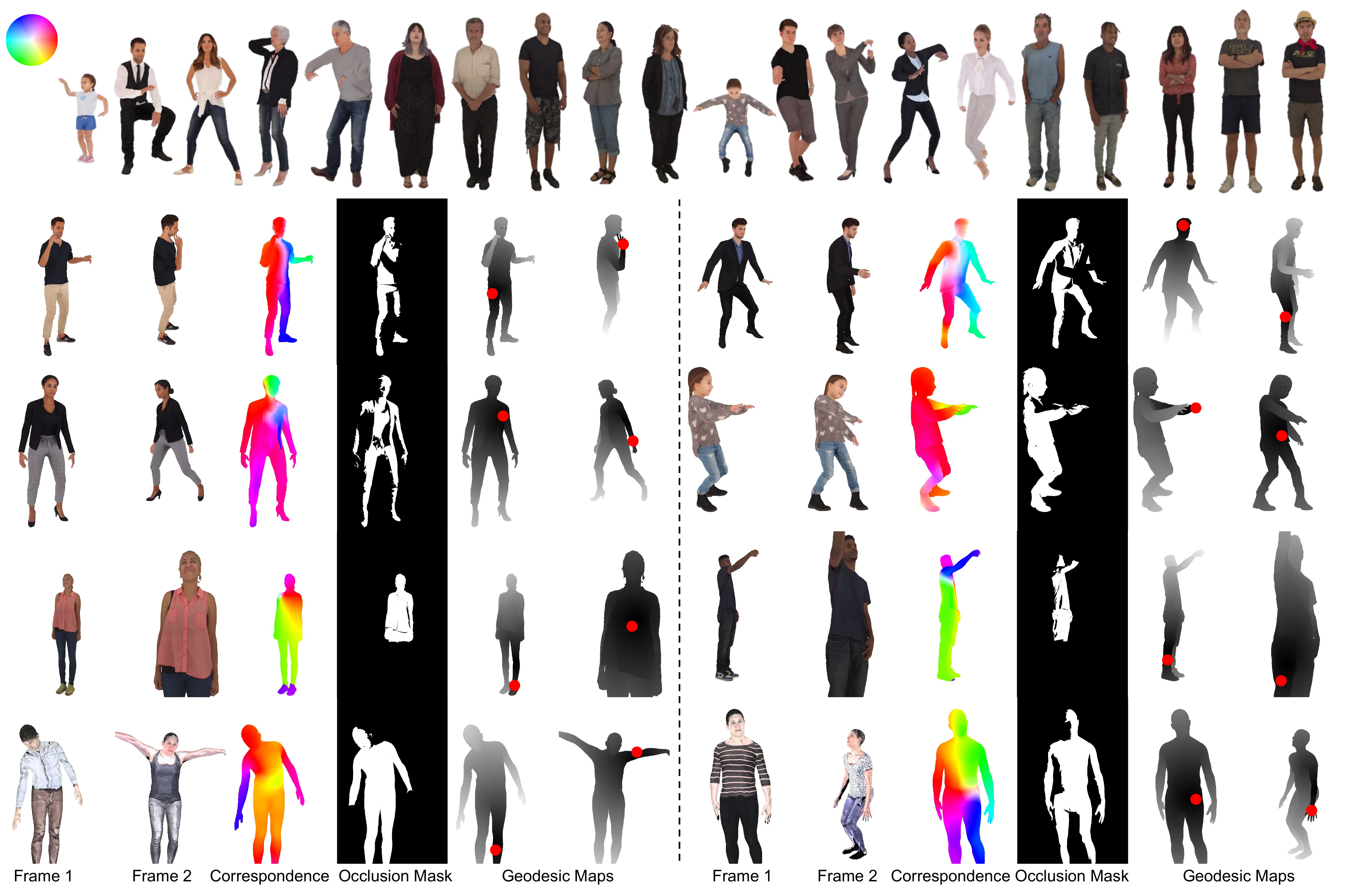}
    \caption{Examples of our semi-synthetic data. From left to right, we show: a pair of the images, the ground truth correspondences, visibility mask, and the geodesic distance map w.r.t. one pixel (marked in red). Rows 1-3 show intra-subject data, and Row 4 shows inter subject data. Please refer to the color legend on the top left for the correspondence direction and magnitude. }
    \label{fig:dataset}
\end{figure*}

\subsection{3D Assets and Candidate Pose}
We collected 3D assets and candidate body poses from various sources to ensure good diversity and high realism.

\vspace{2mm}
\noindent\textbf{SMPL.}
We use the SMPL body model~\cite{loper2015smpl} and $900$ aligned texture maps from SURREAL~\cite{varol2017learning}. These models are less realistic compared to other sources but provide good diversity.
Following SURREAL~\cite{varol2017learning}, we randomly sample shape parameters from the distribution of CAESAR subjects~\cite{robinette2002civilian}, and collect pose parameters by fitting SMPL model using MoSh~\cite{loper2014mosh} to the motion capture data from CMU MoCap database~\cite{CMUMoCap} which contains 2.6K sequences from 23 high-level action categories.

\vspace{2mm}
\noindent\textbf{RenderPeople.}
Additionally, we acquired $25$ rigged 3D human scans from RenderPeople~\cite{renderpeople}, whose models contains different clothing and hair styles, and the texture map are much more detailed than SURREAL.
For each human scans, we animate them using pose sequence collected from Mixamo~\cite{Mixamo}, which includes 27K different poses.

\vspace{2mm}
\noindent\textbf{The Relightables~\cite{guo2019relightables}.}
We also use $60$ high-fidelity posed human models captured using The Relightables~\cite{guo2019relightables} system, since their renderings have higher photorealism compared to the other sources. To minimize rendering artifacts, we do not animate these models and keep them in the original configuration.

\subsection{Camera Setup and Pose Selection}
We use the pinhole camera model with a resolution of $256\times384$ and a focal length of $500$ pixels.
We randomly sample a pair of camera centers in range [1.5, 3.6] meters away from the person, and control the angle between the camera facing directions no large than $60$ degree to produce reasonable overlaps between views.
Note this is already much larger camera variations compared to typical optical flow datasets~\cite{humanflow, ranjan2020learning}.
For the pose, we randomly sample from the pose pool for SMPL and RenderPeople, respectively and, as mentioned above, we fixed the poses for The Relightables scans.

\subsection{Ground truth}
We generate three kinds of ground truths measurements: 1) 2D correspondences, 2) visibility mask, 3) geodesic distance between pixels.
We first render warping fields from one camera to the UV space and from UV space to the other camera.
The correspondences and visibility mask between two rendered image can be obtained by cascading two warping operations.

For geodesic distance on 3D surfaces, we adopt the exact method proposed by Mitchell \etal \cite{mitchell1987discrete}. In order to represent surface in a 2D image, we render a triangle index image and a barycentric coordinates image so that each pixel corresponds to a point in the piece-wise linear surface. Given a pixel as the source, we compute geodesic distances to all of the rest pixels in parallel and store them in a distance image to support our novel dense geodesic loss.

In total, we generate $280K/2.7K$, $795K/2.9K$, $42K/1.8K$ for training/testing splits from SMPL \cite{SMPL:2015}, RenderePeople \cite{renderpeople}, and The Relightables \cite{guo2019relightables} respectively.
Note that inter-subject ground truth correspondences are not available using 3D assets from either RenderePeople \cite{renderpeople} or The Relightables \cite{guo2019relightables}, since it is non-trivial to align high-fidelity 3D scans from different subjects.
Therefore, we generate $2.2$K cross-subject images from SMPL \cite{SMPL:2015} for inter-subject evaluation purposes only (i.e. no training). Indeed, we tried to add some inter-subject data from SMPL into the training stage, but we found that it did not significantly improve the performances on test cases.

\section{Network Architecture}
In this section, we introduce our detailed network architecture, including the feature extractor mentioned in main paper Section 3.1 and used in Section 4.2. We also describe how to integrate it with end-to-end architectures for optical flow and DensePose \cite{guler2018densepose}.

\subsection{Feature Extractor}
The architecture of our feature extractor is shown in~\Autoref{fig:feature_extractor}.
It is 7-level residual U-Net with skip connections.
We use residual block to extract feature, and the feature channels of each level are set as $16$, $32$, $64$, $96$, $128$, $128$, $196$.
In the decoder, bilinear sampling is applied to increase the spatial resolution, and we add a $1$ $\times$ $1$ convolution layer followed by a normalization layer after each residual block to produce HumanGPS feature for each level.

\begin{figure*}
    \centering
    \includegraphics[width=\linewidth]{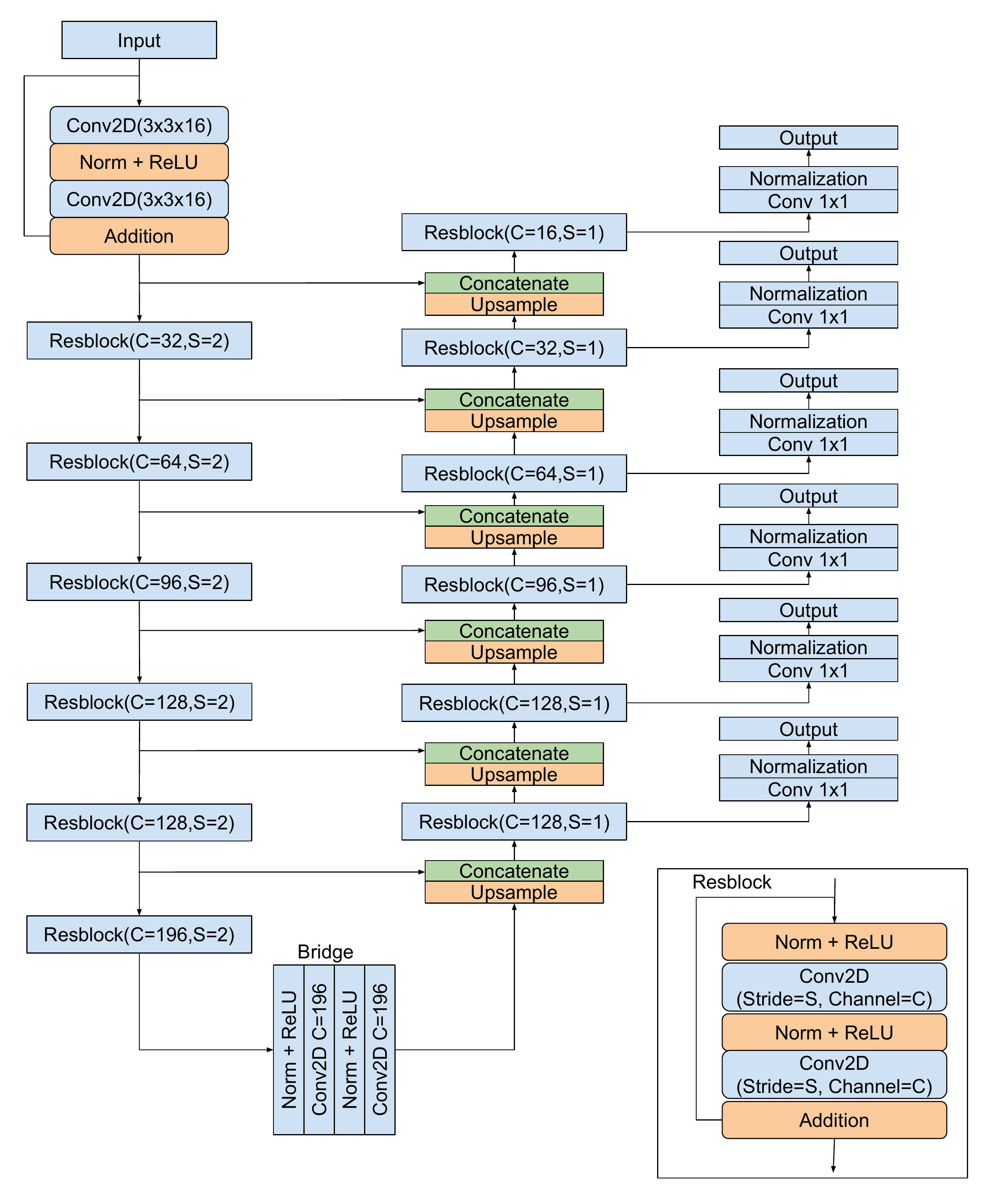}
    \caption{Proposed architecture. Our method relies on a U-Net, with multiple ResNet blocks and skip connections. C and S are the channel number and stride for the convolution layers.}
    \label{fig:feature_extractor}
    \vspace{0.30in}
\end{figure*}

\subsection{PWC-Net + GPS}
As shown in \Autoref{fig:pwc-net}, we attach our HumanGPS feature extractor along with the original feature extractor of PWC-Net. For each level we fuse the features from both feature extractors to obtain the input to the cost volume module. When fusing the feature, HumanGPS feature is passed to two $1$ $\times$ $1$ convolution layers with ReLU as activation function, then the original flow feature and HumanGPS feature are fused by element-wise mean.

\subsection{RAFT + GPS}
\Autoref{fig:raft} shows the architecture of RAFT$+$GPS. Similar to PWC-Net$+$GPS, a HumanGPS feature extractor is added to the original RAFT framework.
The feature from the original feature extractor and our HumanGPS feature extractor are fused before constructing 4D cost volume. Unlike PWC-Net which computes the cost volume in a pyramid, RAFT constructs 4D cost volume at 1/8 resolution, thus we resize the HumanGPS feature map to 1/8 resolution via a stride convolution, then pass it through two $1$ $\times$ $1$ convoluton layers. Then two feature maps are fused by element-wise mean.

\subsection{DensePose + GPS}
We use DensePose \cite{guler2018densepose,neverova2019slim} backbones to extract our GPS feature.
To extend our method for UV coordinates regression, we first take the feature from the second last convolution layer; feed it into a normalization layer, and train the whole network with our loss.
We then feed the feature before the normalization layer into two fully connected layers with $128$ channels, and a regressor to predict the part probability and UV coordinates in each of the $24$ parts respectively.

\begin{figure*}
    \centering
    \includegraphics[width=\linewidth]{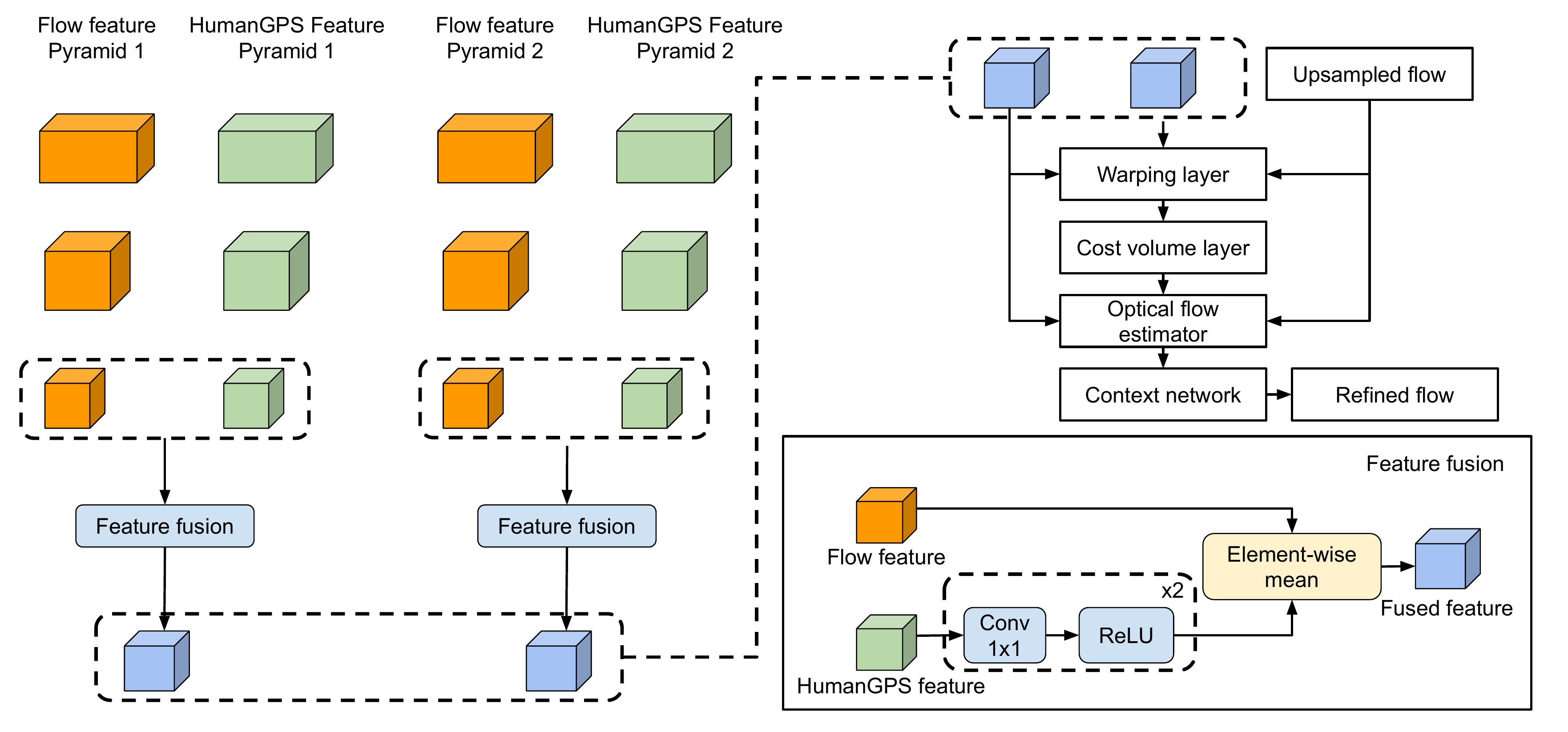}
    \caption{Proposed end-to-end architecture for optical flow. We fuse our GPS feature with the original feature extractor from PWC-Net \cite{sun2018pwc}. As showed in the main paper, this substantially improves the performance even when compared to a PWC-Net with a larger capacity.}
    \label{fig:pwc-net}
    \vspace{0.20in}
\end{figure*}

\begin{figure*}
    \centering
    \includegraphics[width=\linewidth]{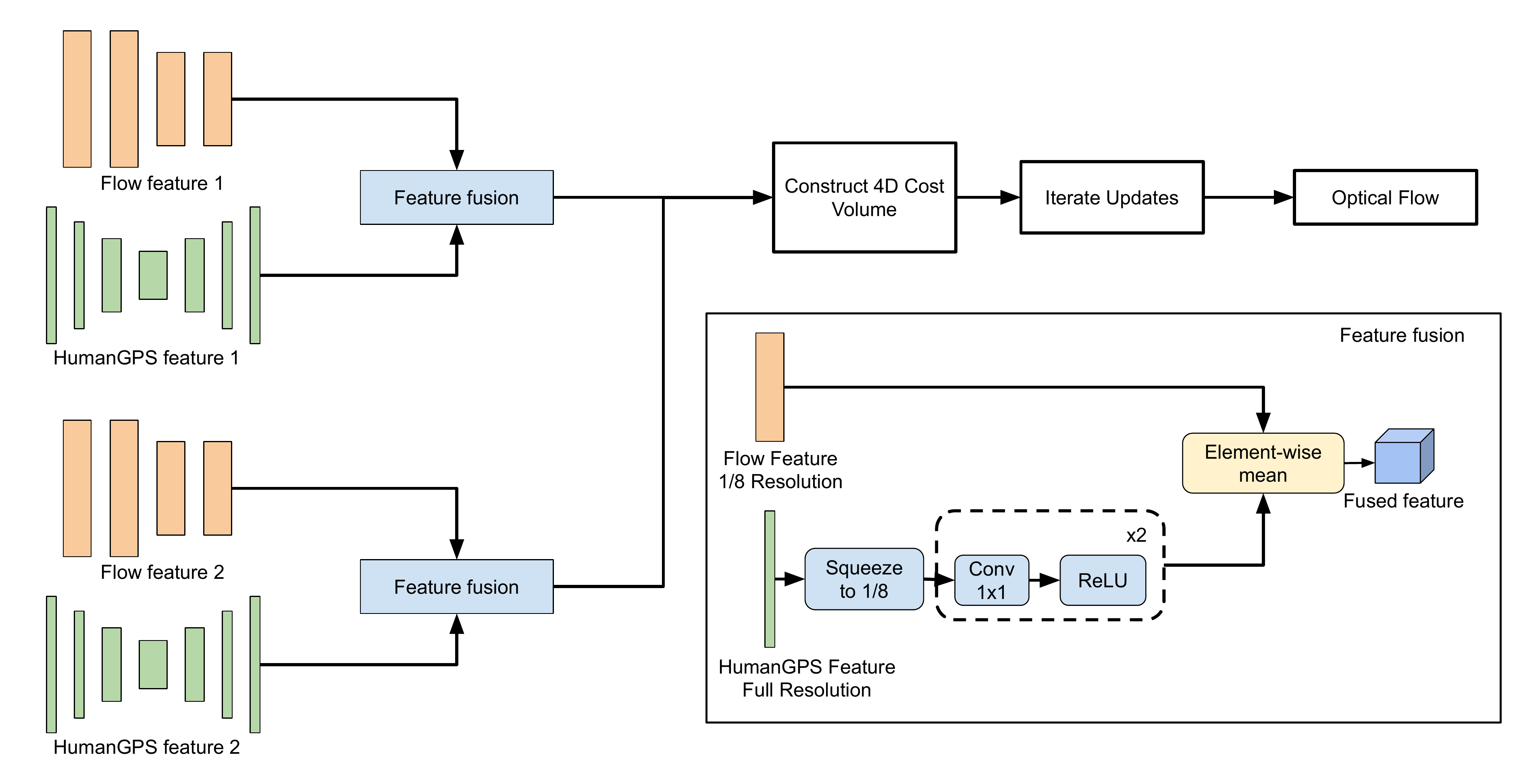}
    \caption{Proposed end-to-end architecture for optical flow. We fuse our GPS feature with the original feature extractor from RAFT \cite{teed2020raft}. As showed in the main paper, this substantially improves the performance even when compared to a RAFT with a larger capacity.}
    \label{fig:raft}
    \vspace{0.20in}
\end{figure*}

\begin{table*}[t]
\centering
\begin{tabular}{l||cc|cc|cc||cc}
\hline 
\multirow{3}{*}{\qquad \qquad Methods} & \multicolumn{6}{c||}{Intra-Subject} & \multicolumn{2}{c}{Inter-Subject}\tabularnewline
\cline{2-9} 
 & \multicolumn{2}{c|}{ SMPL \cite{loper2015smpl} } & \multicolumn{2}{c|}{Relightables \cite{guo2019relightables}} & \multicolumn{2}{c||}{RenderPeople \cite{renderpeople}} & \multicolumn{2}{c}{SMPL \cite{loper2015smpl}}\tabularnewline
\cline{2-9}
 & non & all & non & all & non & all & non & all\tabularnewline
\hline 
Ours + triplet & 9.14 & 24.34 & 13.18 & 25.59 & 16.84 & 29.80 & 21.08 & 30.75\tabularnewline
Ours + classify & 9.73 & 25.80 & 15.97 & 33.03 & 18.33 & 34.03 & 11.21 & 25.72\tabularnewline
Ours + $L_c$ + $L_s$   & 8.17 & 19.31 & 14.61 & 21.45 & 14.51 & 24.21 & 12.02 & 24.51\tabularnewline
Ours + $L_c$ + $L_d$  & 7.67 & 18.72 & 12.46 & 20.58 & 13.21 & 23.22 & 9.99 & 19.85\tabularnewline
Ours + $L_c$ + $L_s$ + $L_d$ & 7.50 & 18.00 & 12.24 & 19.30 & 12.41 & 22.73 & 9.19 & 18.61\tabularnewline
Ours + $L_c$ + $L_d$ + $L_{cd}$ & 7.53 & 18.17 & 12.12 & 19.09 & 12.38 & 22.99 & 9.27 & 19.23\tabularnewline
Ours + Full & 7.32 & 17.57 & 11.50 & 19.12 & 12.29 & 22.48 & 8.57 & \textbf{17.87}\tabularnewline
Ours + Full + Multi-scale & \textbf{7.12} & \textbf{17.51} & \textbf{11.24} & \textbf{18.95} & \textbf{11.91} & \textbf{22.12} & \textbf{8.49} & 17.99\tabularnewline
\hline 
feature=8 & {9.34} & {19.77} & {14.25} & {20.62} & {14.68} & {24.46} & {11.81} & {20.43}\tabularnewline
feature=16 (Ours) & {7.12} & {17.51} & {11.24} & {18.95} & {11.91} & {22.12} & {8.49} & 17.99\tabularnewline
feature=32 & {6.87} & {17.45} & {10.96} & \textbf{18.77} & {11.64} & {22.05} & \textbf{8.44} & \textbf{17.73}\tabularnewline
feature=64 & \textbf{6.78} & \textbf{16.93} & \textbf{10.83} & 18.82 & \textbf{11.58} & \textbf{21.99} & {8.63} & {18.13}\tabularnewline
\hline 

\end{tabular}
\vspace{1mm}
\caption{Quantitative evaluation for correspondences search. We report the average end-point-error (EPE) of non-occluded (marked as non) and all pixels (marked as all) on four test sets created from different sources of 3D assets. We report the results for both intra and inter-subjects.}
\label{tab:aepe}
\vspace{-0.15in}
\end{table*}

\begin{figure*}
    \centering
    \includegraphics[width=\linewidth]{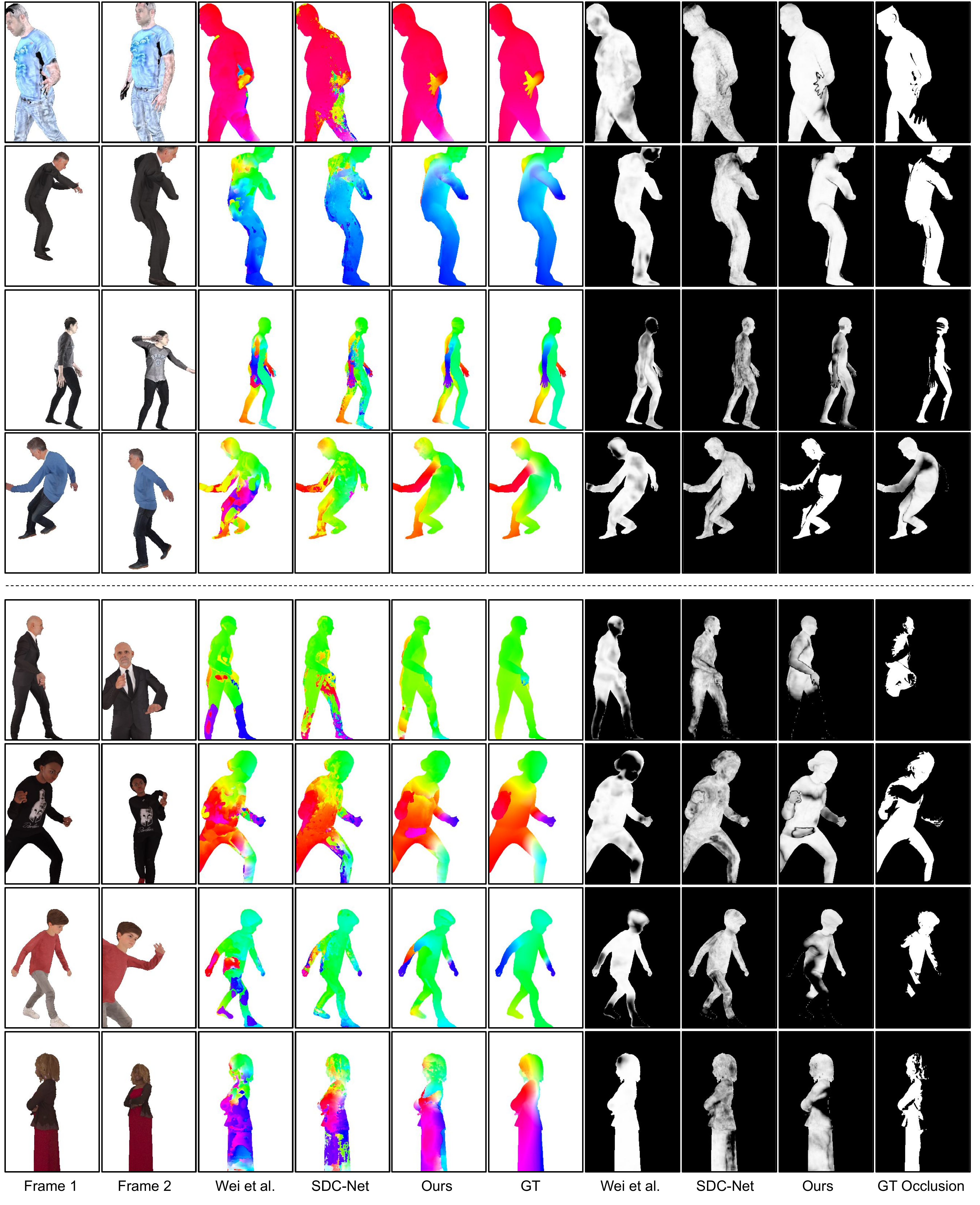}
    \caption{Comparison on intra-subject data. We compare to SDC-Net \cite{schuster2019sdc} and Wei \etal \cite{wei2016dense}. Our method shows consistently better performance on both correspondence (left) and occlusion detection (right). The top and bottom are sampled from the 20\% of the test cases with the smallest and largest error respectively.}
    \label{fig:intra_search_correspondence}
\end{figure*}

\begin{figure*}
    \centering
     \includegraphics[width=\linewidth]{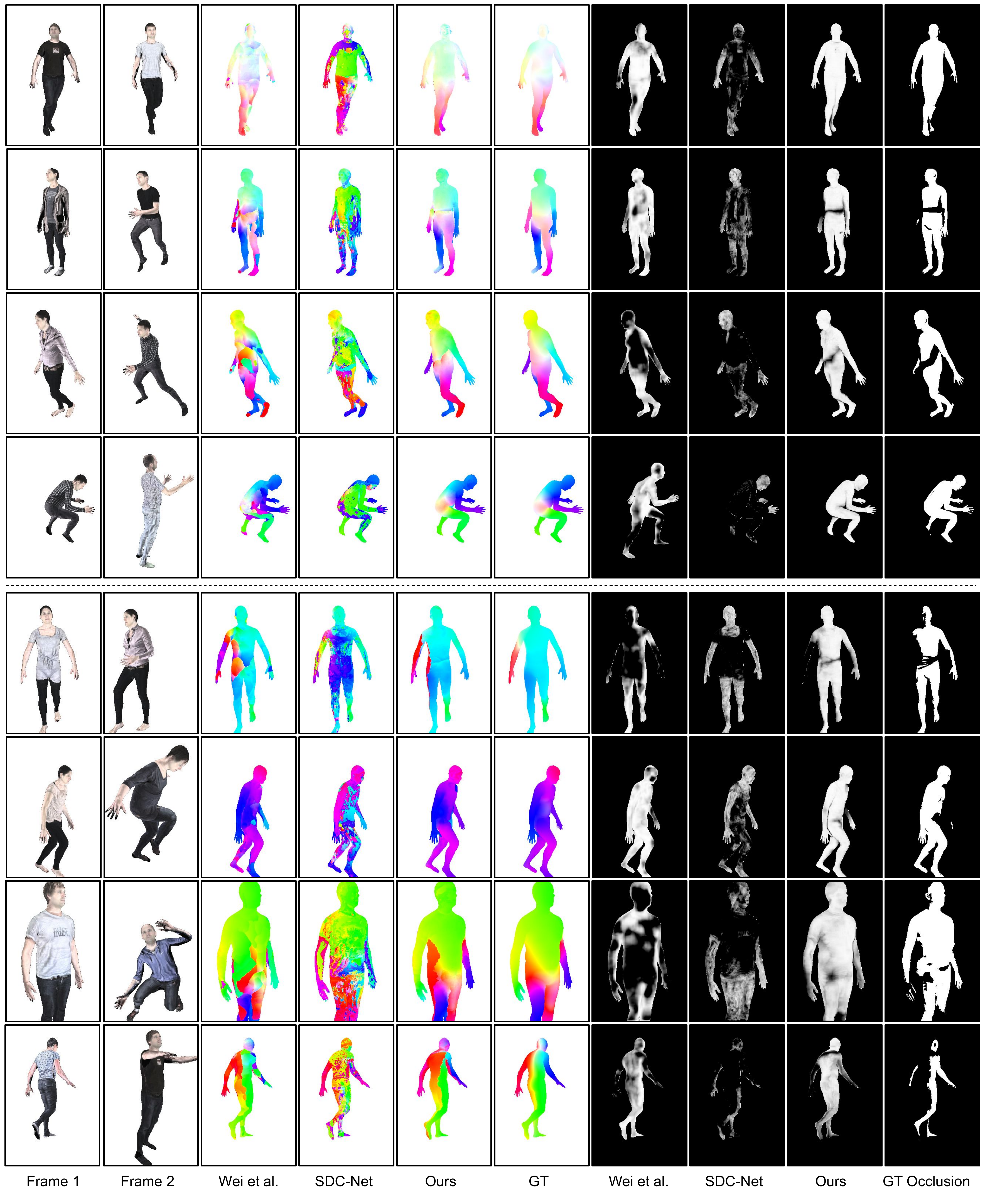}
    \caption{Comparison on inter-subject data. We compare to SDC-Net \cite{schuster2019sdc} and Wei \etal \cite{wei2016dense}. Our method shows consistently better performance on both correspondence (left) and occlusion detection (right). The top and bottom are sampled from the 20\% of the test cases with the smallest and largest error respectively.}
    \label{fig:inter_search_correspondence}
\end{figure*}

\section{Additional Evaluations}
In this section, we show more experimental results to evaluate our Human GPS feature for dense human correspondence.

\subsection{Correspondence and Visibility Map}

\Autoref{fig:intra_search_correspondence}, \Autoref{fig:inter_search_correspondence} show more examples of our predicted correspondences and visibility maps for intra and inter-subject cases respectively.
In each figure, we sort all the test cases according to the error metric of our method and randomly pick four from the top and bottom 20\% respectively.
This gives a demonstration of the full spectrum of the quality on the test set.
Our method works consistently well on both easy (the top four) and hard (the bottom four) cases, and outperforms other methods \cite{wei2016dense,schuster2019sdc} on the visual quality of both the predicted correspondences and visibility maps.
There are some regions where all the methods performs equally poorly, however these are mostly occluded regions as shown in the visibility map.
Depending on the application, it might be more preferred to mark these pixels as no available matching rather than hallucinate continuous and implausible correspondences.

\subsection{Warping via Correspondence}
\label{sec:warping}
To qualitatively evaluate the correspondences, we show in \Autoref{fig:warping} the warping results of frame 1 using the texture of the frame 2, leveraging the predicted correspondence field.
Our method produces more visually appealing and semantically correct warping results compared to DensePose \cite{guler2018densepose}, where we used the predicted UV coordinates to establish correspondences. 

\subsection{Additional Ablation Study}
\label{sec:ablation}
In paper Section 4.3, we studied the effect of each loss term for learning the GPS feature by gradually adding $L_s, L_d, L_{cd}$. In~\Autoref{tab:aepe}, we provide more quantitative evaluations on using other combinations of loss terms in training the models, which consistently support that all the loss terms
contributes the learning of GPS feature. It should be noted, the behavior of our losses is substantially different from that of a triplet loss. While the triplet loss penalizes all the non-matching pixels equally (i.e., further
apart compared to the matched pixel), the dense geodesic
loss, instead, pushes features between non-matching pixels
apart proportionally to the surface geodesic distance with
respect to the reference pixel. The geodesic distance provides important supervision for network to learn the affinity, i.e., likelihood of matching, between pixels, and hence
yields smooth and discriminative feature space.  This is also
empirically demonstrated in~\Autoref{tab:aepe} between (Ours+triplet) and (Ours+$L_d$+$L_d$), which shows using consistency + dense geodesic loss achieves better results than using triplet loss. In~\Autoref{tab:aepe} bottom part, we also report the impact of feature number on the correspondence accuracy. The performance is mostly saturated at feature number=16, though slightly better performance can be achieved with more computation.

\begin{figure*}
    \centering
    \includegraphics[width=18cm]{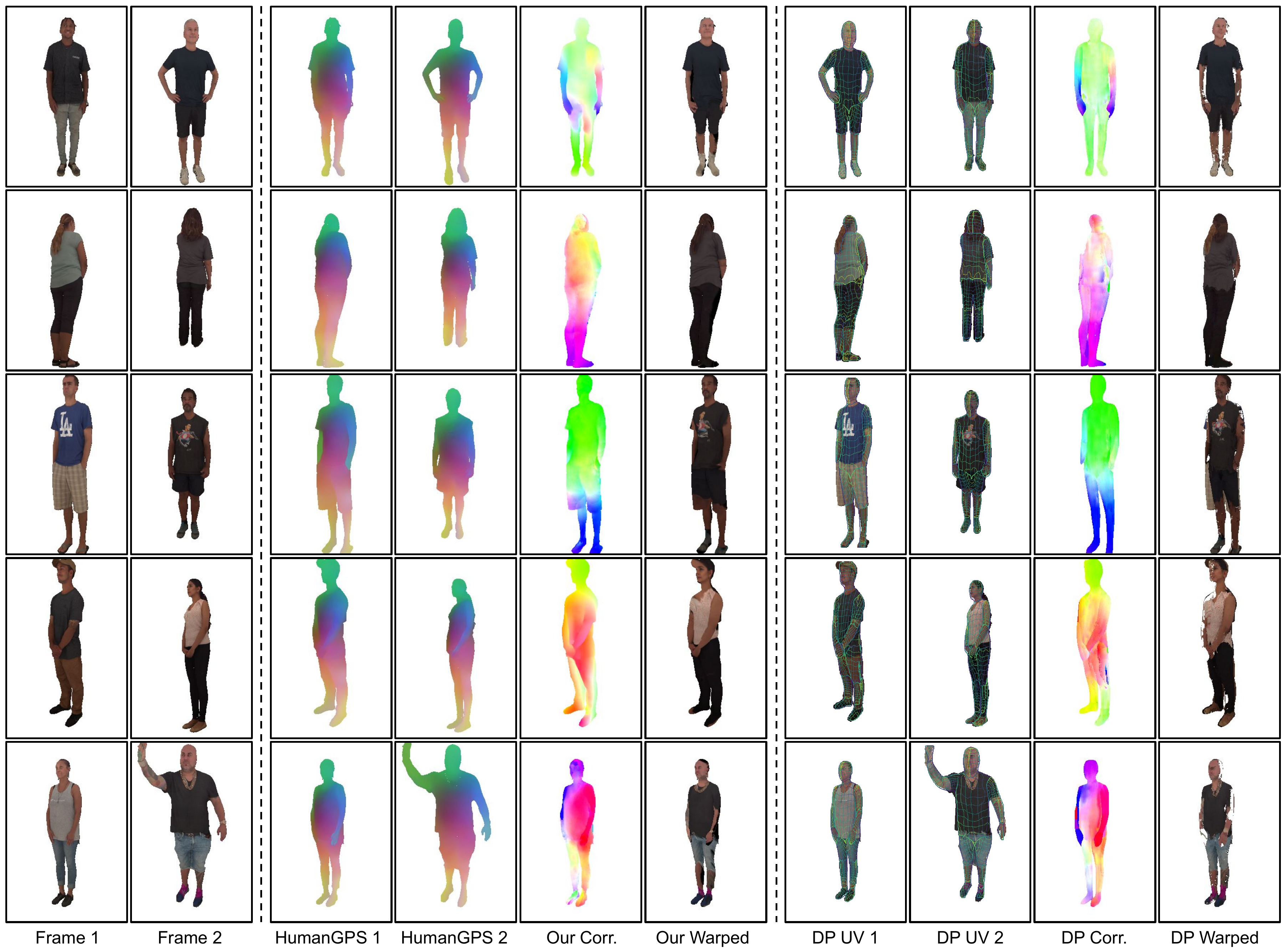}
    \caption{Cross-subject warping results. The left section shows two reference frames. The middle section shows our GPS feature, correspondences, and the warped result of frame 1 using the texture of frame 2. The right section shows the results of DensePose (DP) \cite{guler2018densepose}.}
    \label{fig:warping}
\end{figure*}

\subsection{Real-world Image Results}
We also evaluate our HumanGPS feature
on sparse 2D joint ground truth annotated in Human3.6M~\cite{ionescu2013human3}. Specifically, we build correspondences across video frames using extracted HumanGPS features and measure the average end-point-error (AEPE) to the ground-truth correspondence between sparse 2D joints.
The AEPE of SDC-Net, Wei et al. and
ours is 19.72, 29.50 and 14.33. Our method consistently
outperforms the previous methods in real data. 
\Autoref{fig:real_image_results} shows more results of our method on real images.
Although we use semi-synthetic data for training, our method generalizes well onto various real images in the wild.
It worth mentioning that, often, the foreground computation using off-the-shelf segmentation algorithms~\cite{deeplabv3} may not be accurate, nevertheless, our method is robust against minor errors in practice. Given our
exhaustive evaluation, we expect these results to hold true
for generic datasets.

\begin{figure*}
    \centering
    \includegraphics[width=18cm]{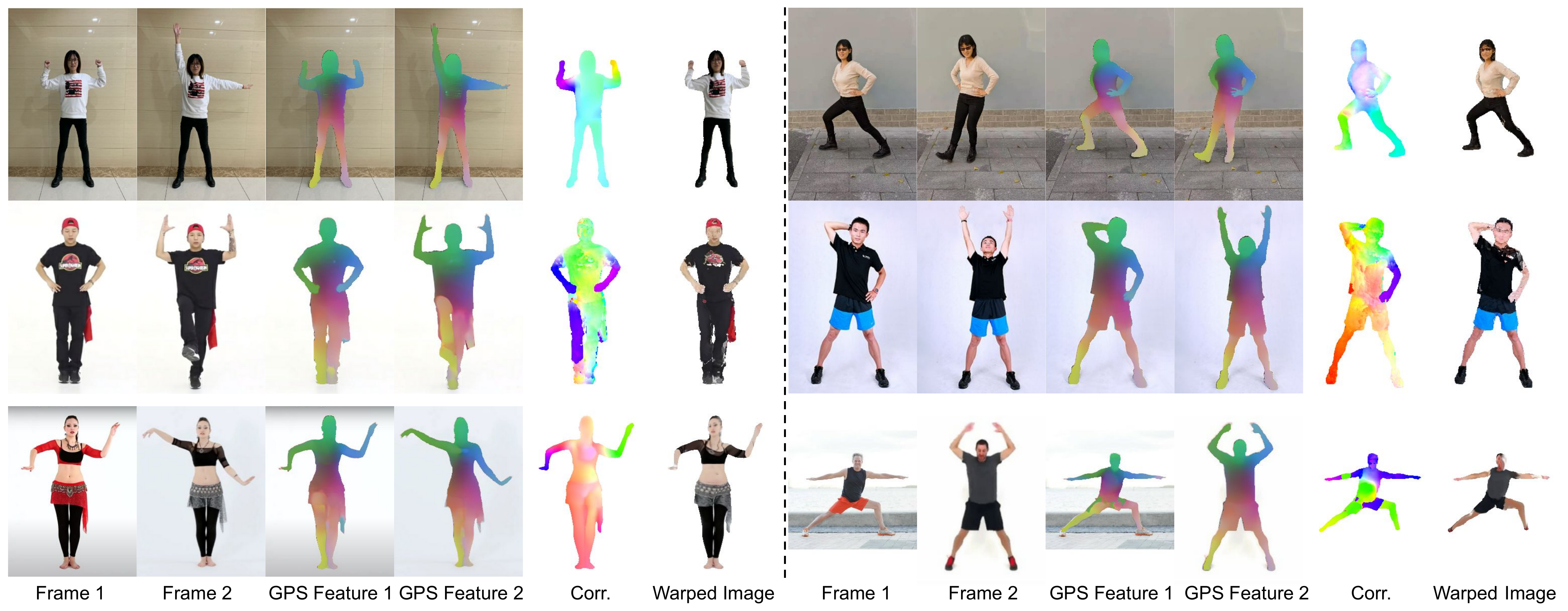}
    \caption{In-the-wild results. For each pair of images, we show the GPS feature, the established correspondences, and the warped result.}
    \label{fig:real_image_results}
\end{figure*}

\begin{table*}[t]
\centering
\begin{tabular}{l||c|c|c||c}
\hline 
\multirow{2}{*}{\qquad Methods} & \multicolumn{3}{c||}{Intra-Subject} & Inter-Subject\tabularnewline
\cline{2-5} \cline{3-5} \cline{4-5} \cline{5-5} 
 & SMPL \cite{loper2015smpl} & The Relightables \cite{guo2019relightables} & RenderPeople \cite{renderpeople} & SMPL \cite{loper2015smpl}\tabularnewline
\hline 
SDC-Net \cite{schuster2019sdc} & 56.40 & 48.17 & 58.38 & 28.98\tabularnewline
Wei \etal \cite{wei2016dense} & 32.20 & 25.78 & 34.13 & 32.25\tabularnewline
Ours & \textbf{71.20} & \textbf{56.08} & \textbf{67.65} & \textbf{69.33}\tabularnewline
\hline 
PWC-Net \cite{sun2018pwc} & 90.25 & 85.16 & 87.06 & 72.18\tabularnewline
PWC-Net{*} & 92.20 & 85.33 & 89.32 & 63.06\tabularnewline
PWC-Net + GPS & \textbf{94.93} & \textbf{87.67} & \textbf{91.38} & \textbf{80.91}\tabularnewline
\hline 
\end{tabular}
\caption{Quantitative evaluation of occlusion detection. We show the average precision for the occlusion detection on three intra-subject test sets and one inter-subject test set. Methods on the top half directly use feature distance for occlusion detection (see the main paper Section 4.2 for details), and methods in the bottom half use optical flow architecture to regress the occlusion mask.
Our feature shows good skill in occlusion detection directly via feature distance, and further improves PWC-Net on this task.
Please see the main paper for explanation of the model with $^*$.}
\label{tab:occlusion_detection_ap}
\vspace{-0.1in}
\end{table*}

\begin{table*}[htb]
\center
\begin{tabular}{l||ccc|ccc}
\hline 
\multirow{2}{*}{\qquad Architectures} & \multicolumn{3}{c|}{DensePose \cite{guler2018densepose,neverova2019slim}} & \multicolumn{3}{c}{HumanGPS}\tabularnewline
\cline{2-7} \cline{3-7} \cline{4-7} \cline{5-7} \cline{6-7} \cline{7-7} 
 & 5cm & 10cm & 20cm & 5cm & 10cm & 20cm\tabularnewline
\hline 
ResNet-101 FCN \cite{guler2018densepose} & 43.05 & 65.23 & 74.17 & 49.09 & 73.12 & 84.51\tabularnewline
ResNet-101 FCN{*} \cite{guler2018densepose} & 51.32 & 75.50 & 85.76 & 53.01 & 76.77 & 87.18\tabularnewline
HG Stack-1 \cite{neverova2019slim} & \multicolumn{1}{c}{49.89} & 74.04 & 82.98 & 50.50 & 75.57 & 87.18\tabularnewline
HG Stack-2 \cite{neverova2019slim}& 52.23 & 76.50 & 84.99 & 52.91 & 77.21 & 88.50\tabularnewline
HG Stack-8 \cite{neverova2019slim}& 56.04 & 79.63 & 87.55 & 55.41 & 79.76 & 89.44\tabularnewline
\hline 
\end{tabular}
\vspace{1mm}
\caption{Quantitative evaluation for dense human pose regression on DensePose COCO dataset \cite{guler2018densepose}. Following previous work \cite{guler2018densepose}, we assume ground truth bounding box is given and calculate percentage of pixels with error smaller than thresholds. All the models are trained on images with background, except the one marked with $^*$, which is trained on image with white background following DensePose \cite{guler2018densepose} for comparison. }
\label{tab:densepose_single_metric}
\end{table*}

\subsection{Occlusion Detection}
In the main paper Section 4.2, we show the qualitative results of our occlusion detection.
Here we quantitatively evaluate the occlusion detection, following standard evaluation protocol adopted by object detection~\cite{girshick2014rich}.
We detect occluded pixel as the set of pixels with the visibility score under a threshold. By varying a threshold on the distances, we calculate precision (i.e. percentage of predicted pixels that are truly occluded) and recall (i.e. percentage of occluded pixels that are detected). Finally we report the average precision as the area under the precision-recall curve.

\Autoref{tab:occlusion_detection_ap} (Top) shows the comparison to other feature descriptor methods \cite{wei2016dense,schuster2019sdc}.
SDC-Net \cite{schuster2019sdc} shows better occlusion detection performance, while Wei \etal \cite{wei2016dense} generalize better to inter-subject data.
Overall, our method performs the best over all intra and inter-subject test sets.

In \Autoref{tab:occlusion_detection_ap} (Bottom), we also show the performance of occlusion detection from neural network architecture designed for optical flow.
Taking PWC-Net \cite{sun2018pwc} as example, integrating our HumanGPS feature achieves the best average precision compared to the original PWC-Net even with the augmented encoder. Please check main paper Section 4.5 for explanation of the $^*$ version.

\subsection{Evaluation on Human Optical Flow Dataset}
In the main paper Section 4.5, we showed that our method can improve the human correspondences on our test sets when combined with optical flow network.
Here we further evaluate on public human optical flow dataset proposed by Ranjan~\etal~\cite{humanflow,ranjan2020learning}.
Compared with our dataset, they only use SMPL models for data generation, and their camera and pose variations between each pair of images are much smaller than the ones we generated.
Note that optical datasets usually contains only small motion and consider both foreground and background, which are not the focus and strength of our approach.

Similar as the experiment setup of the main paper, we augment PWC-Net \cite{sun2018pwc} with an augmented feature extractor and apply loss function to supply HumanGPS feature.
The average end-point error on Single-Human Optical Flow dataset (SHOF) \cite{ranjan2020learning} is shown in \Autoref{tab:optical_flow}.
The PWC-Net integrated with HumanGPS achieves the best performance compared to original PWC-Net with and without augmented feature extractor.
This indicates that our method not only provide correspondences for large motion, but it is also effective when the small motion assumption holds. This evaluation is done on both foreground and background, which shows it is straightforward to extend our method on full images without the dependency on segmentation methods.


\begin{table}[!hb]
\center
\begin{tabular}{lcc}
\hline 
\qquad Method & Finetune & AEPE\tabularnewline
\hline 
PWC-Net & No & 0.2185\tabularnewline
PWC-Net & Yes & 0.2185\tabularnewline
PWC-Net{*} & Yes & 0.1411\tabularnewline
PWC-Net + GPS & Yes & \textbf{0.1239}\tabularnewline
\hline 
\end{tabular}
\caption{Evaluate on Single-Human Optical Flow dataset (SHOF) \cite{ranjan2020learning}. Our method achieve the best performance over all. Please see the main paper for explanation of the model with $^*$. }
\label{tab:optical_flow}
\end{table}


\subsection{Additional Comparisons with DensePose}
In main paper Section 4.5, we showed that using GPS feature can achieve competitive dense human pose regression performance.
Here we show comparisons using additional network backbones in \Autoref{tab:densepose_single_metric}.
Same as the setup in the main paper Section 4.5, we adopt the same evaluation setup as DensePose \cite{guler2018densepose}, where ground truth bounding box is given; percentages of pixels with geodesic error less than certain thresholds are taken as the metric; and evaluate on DensePose MSCOCO benchmark \cite{guler2018densepose}.
Directly regressing UV from our GPS feature using only 2 layers of MLP consistently achieves competitive performance compared to previous work using similar backbone \cite{guler2018densepose,neverova2019slim}, which indicates the effectiveness of our feature in telling cross-subject correspondences.
We also evaluate parametric model fitting based methods \cite{bogo2016keep}.
Their errors are $20.73$, $40.05$, $54.23$ for $5$cm, $10$cm, and $20$cm respectively, which is much worse than our method.


\begin{figure}[htb]
\center
\includegraphics[width=8.5cm]{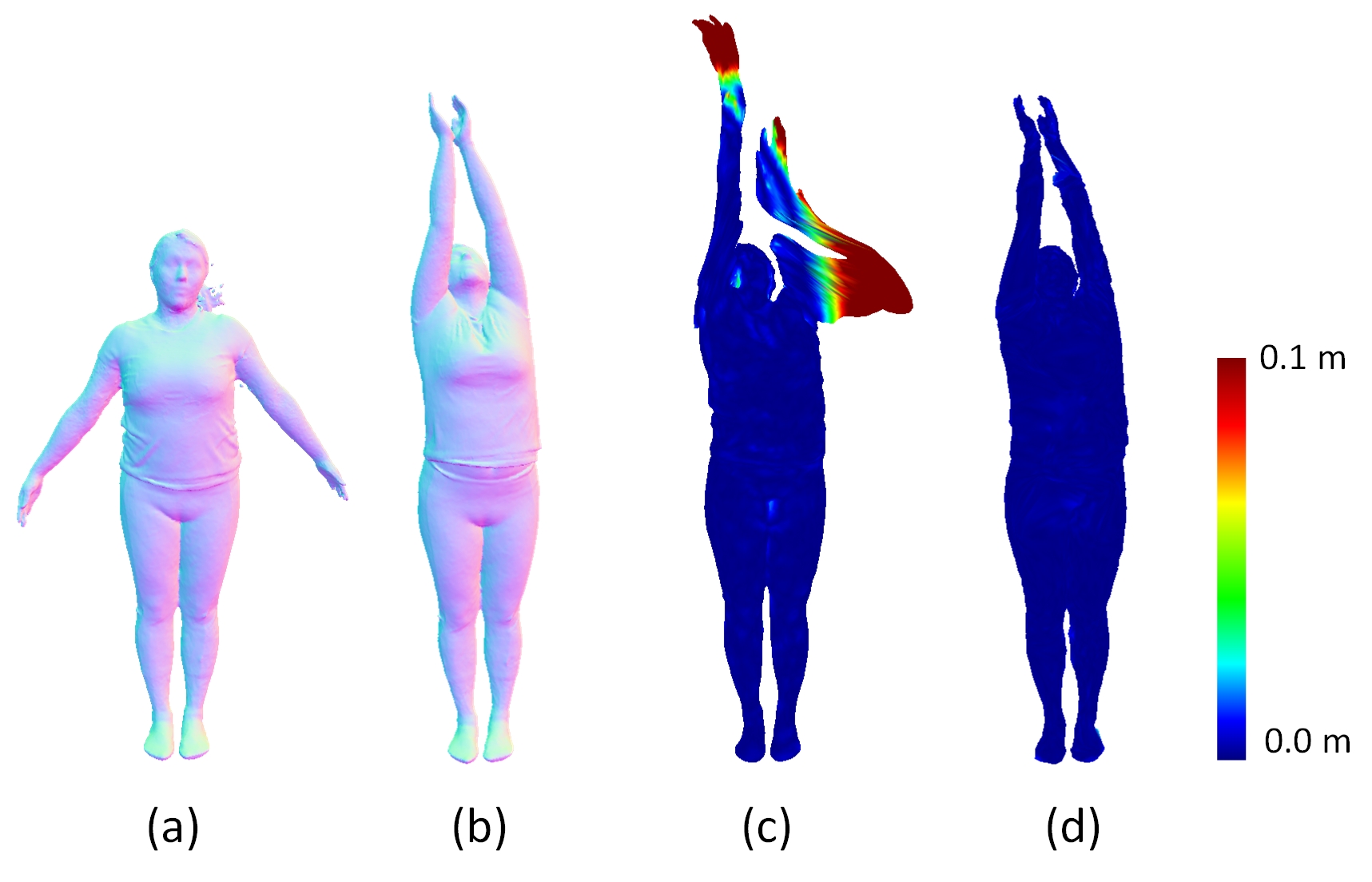}
\caption{Quantitative comparison of non-rigid tracking with learned correspondences. (a) reference geometry; (b) target geometry; non-rigid alignment without (c) and with (d) our learned correspondences. Surface errors are coded as per-pixel colors.}
\label{fig:nonrigid_tracking_numeric_comparison}
\end{figure}

\section{Applications}
In this section, we show how our dense human correspondence benefits various applications.

\subsection{Nonrigid Tracking and Fusion}
\begin{figure}[htb]
\center
\includegraphics[width=8cm]{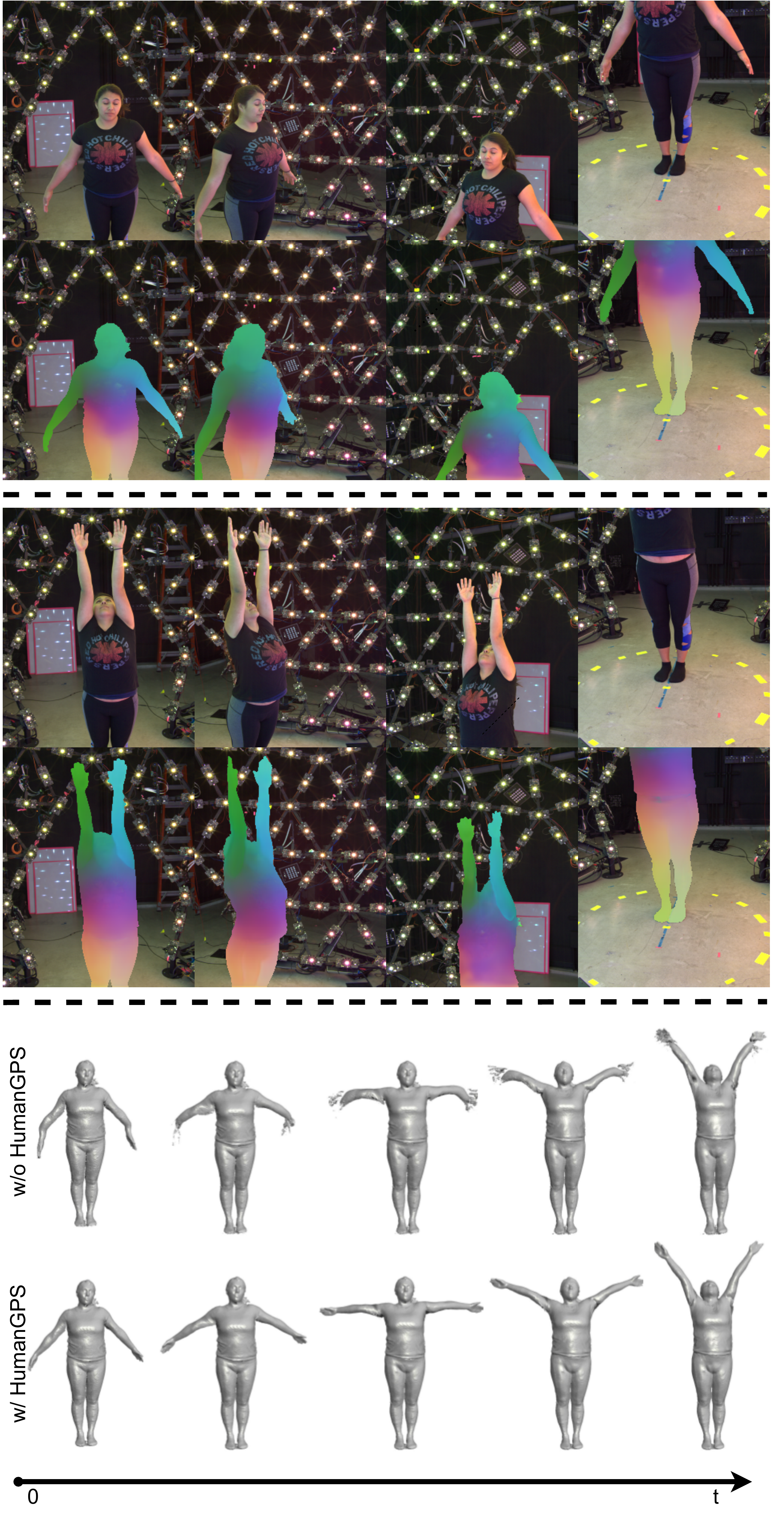}
\caption{Nonrigid Fusion Comparison. We improve the non-rigid tracking using correspondence extracted via our HumanGPS feature from the color image (top and middle). In the bottom, we show the fusion results without and with our correspondence. The standard dynamic fusion system fails quickly under fast motion, whereas successfully tracks the deformation with our correspondences.
}
\label{fig:nonrigid_tracking_comparison}
\end{figure}

Existing nonrigid tracking/fusion systems~\cite{newcombe2015dynamicfusion,dou2016fusion4d,dou2017motion2fusion} have challenges when tracking fast motions.
Such a system typically employs the ICP alike method, and it requires a good initialization on the non-rigid deformation parameters to extract reliable point to point correspondences, which are in turn used to refine the deformation in an iterative manner. 
Human body movements such as waving arms would easily break above requirement when performed too fast. Whereas high speed cameras \cite{need4speed,twinfusion} could mitigate this behavior, here we show that HumanGPS is also an effective way to improve the results without the need of custom hardware.

The correspondences from GPS feature on color images provides additional constraints for the nonrigid tracking system, and it helps to rescue ICP failures.
To demonstrate that, we provide correspondence built across $6$ color images as additional initialization for the nonrigid deformation along with the ICP. The tracking algorithm takes a reference mesh and deforms it non-rigidly to align with a sequence of the target geometry. As shown in \Autoref{fig:nonrigid_tracking_numeric_comparison}, the final target geometry (b) demonstrates large pose difference from the reference (a). Traditional non-rigid tracking algorithms (c) fail in such a case, while such large deformation can be correctly estimated benefiting from our learned correspondences (d).

In addition, non-rigid fusion algorithms such as Motion2Fusion~\cite{dou2017motion2fusion} improve reconstruction accuracy with learned correspondences. This algorithm takes $6$ depth images from different view point and runs non-rigid warping between canonical and live frames. The warping function is solved with additional constraints of learned correspondences from each view point and non-rigid motion is estimated with more accuracy. As shown in \Autoref{fig:nonrigid_tracking_comparison}, the standard dynamic fusion system fails quickly under fast motion, whereas successfully track the deformation with our correspondences.

Please see the supplementary webpage for video demo.

\subsection{Morphing}

\begin{figure}[htb]
\center
\includegraphics[width=8cm]{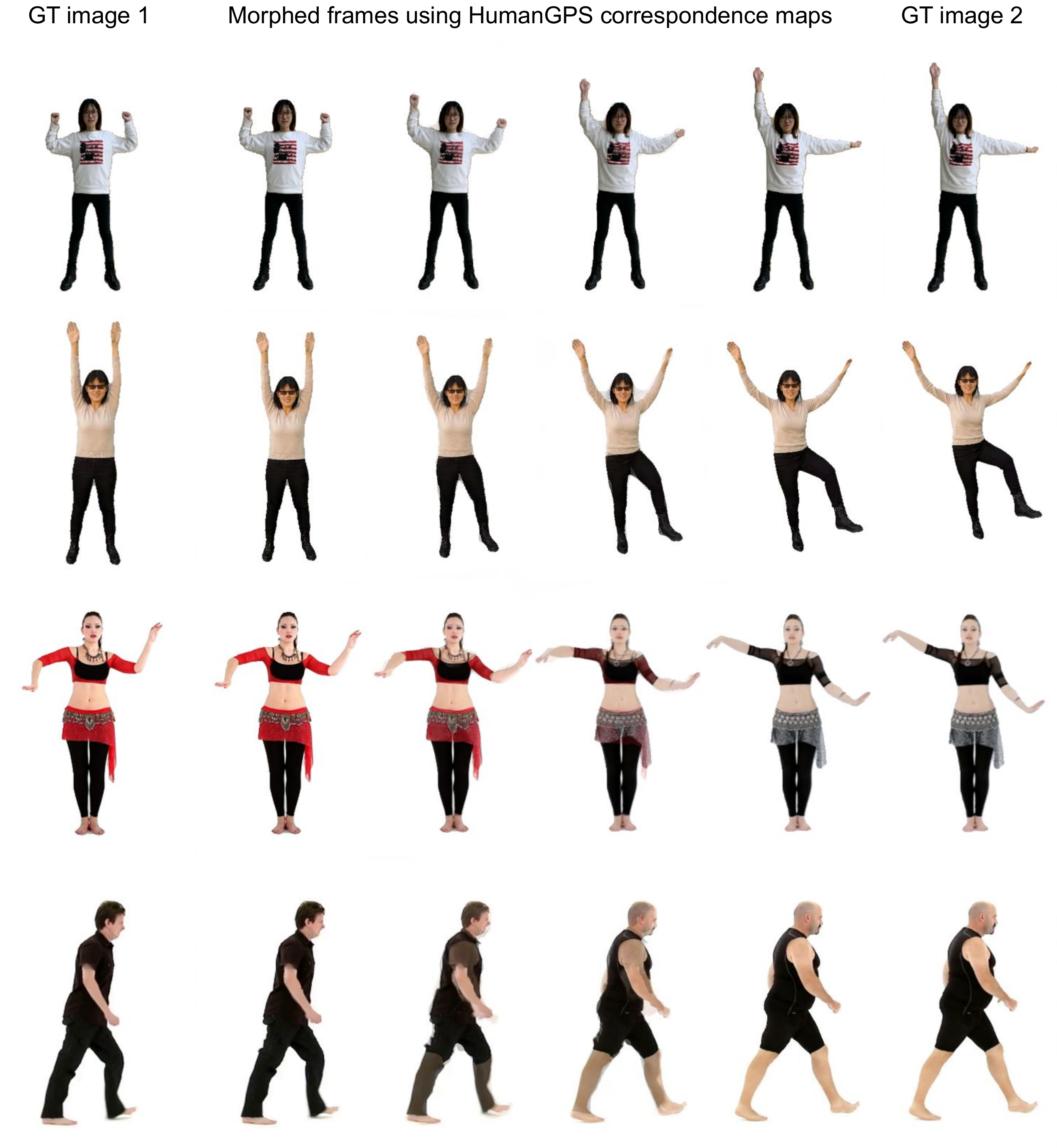}
\caption{Morphing results based on HumanGPS. The leftmost and rightmost columns show two input images of intra- or inter-subjects. We compute the dense correspondence maps and generate the morphed frames in-between. See the supplementary webpage for more results.}
\label{fig:vis_morphing}
\end{figure}

Morphing is a powerful technique to create smooth animation between images. A crucial component to successful image morphing is to create a map that aligns corresponding image elements. \cite{liao2014automating,liao2014semi}. With GPS feature, one can directly establish pixels correspondences to create a smoother video transition between intra- and inter- subjects. Please refer to Fig. \ref{fig:vis_morphing} for example morphed results and the supplementary webpage for more morphing videos.

}
\end{document}